\documentclass[journal]{IEEEtran}
\usepackage{amsmath,amsfonts}
\usepackage{array}
\usepackage[caption=false,font=normalsize,labelfont=sf,textfont=sf]{subfig}
\usepackage{textcomp}
\usepackage{stfloats}
\usepackage{url}
\usepackage{verbatim}
\usepackage{graphicx}
\usepackage{cite} 
\usepackage{bm}
\usepackage{amssymb}
\usepackage{multirow}
\usepackage{setspace}
\usepackage{gensymb}
\usepackage{makecell}
\usepackage[ruled,linesnumbered,vlined]{algorithm2e}
\begin{document}

\title{Activity Detection for Massive Connectivity in Cell-free Networks with Unknown Large-scale Fading, Channel Statistics, Noise Variance, and Activity Probability: A Bayesian Approach}

\author{Hao Zhang, Qingfeng Lin, Yang Li, Lei Cheng, Yik-Chung Wu
\thanks{The work of Y. Li was supported by the National Natural Science Foundation of China under Grant 62101349. The work of L. Cheng was supported in part by the National Natural Science Foundation of China under Grant 62371418, in part by the Fundamental Research Funds for the Central Universities (226-2023-00012), in part by the Zhejiang University Education Foundation Qizhen Scholar Foundation, and in part by Science and Technology on Sonar Laboratory under Grant 6142109KF212204. \textit{(Corresponding author: Yik-Chung Wu.)}}
\thanks{H. Zhang, Q. Lin and Y.-C. Wu are with the Department of Electrical and Electronic Engineering, The University of Hong Kong, Hong Kong (e-mail: haozhang@eee.hku.hk, qflin@eee.hku.hk, ycwu@eee.hku.hk).}
\thanks{Y. Li is with Shenzhen Research Institute of Big Data, Shenzhen 518172, China (e-mail: liyang@sribd.cn).}
\thanks{L. Cheng is with the College of Information Science and Electronic Engineering, Zhejiang University, Hangzhou, 310027, China (e-mail: lei\_cheng@zju.edu.cn).}
\thanks{Part of the results will appear in IEEE ICASSP 2024\cite{icassp2024}.}}


\IEEEpubid{0000--0000/00\$00.00~\copyright~2021 IEEE}

\maketitle

\begin{abstract}
Activity detection is an important task in the next generation grant-free multiple access. While there are a number of existing algorithms designed for this purpose, they mostly require precise information about the network, such as large-scale fading coefficients, small-scale fading channel statistics, noise variance at the access points, and user activity probability. Acquiring these information would take a significant overhead and their estimated values might not be accurate. This problem is even more severe in cell-free networks as there are many of these parameters to be acquired. Therefore, this paper sets out to investigate the activity detection problem without the above-mentioned information. In order to handle so many unknown parameters, this paper employs the Bayesian approach, where the unknown variables are endowed with prior distributions which effectively act as regularizations. Together with the likelihood function, a maximum a posteriori (MAP) estimator and a variational inference algorithm are derived. Extensive simulations demonstrate that the proposed methods, even without the knowledge of these system parameters, perform better than existing state-of-the-art methods, such as covariance-based and approximate message passing methods.
\end{abstract}

\begin{IEEEkeywords}
Bayesian, activity detection, cell-free, massive machine-type communications, grant-free random access.
\end{IEEEkeywords}

\section{Introduction}
\IEEEPARstart{M}{assive} machine-type communications (mMTC) is expected to play a tremendous role in future wireless systems. In an mMTC scenario, there are a large number of potential devices, but only a small portion of them are active with short package transmission at any given time.\par

Due to the sporadic traffic pattern under massive communications, conventional scheduling-based methods that resolve collisions with orthogonal multiple access (OMA) schemes, such as time division multiple access (TDMA) and orthogonal frequency division multiple access (OFDMA)\cite{zhou2005}, are not adequate. To this end, grant-free multiple access schemes have been proposed to reduce latency and signaling overhead\cite{liu2018massive,liu2018massive3}. In grant-free access, a unique non-orthogonal pilot sequence is assigned to each user, and the base station detects which devices are active. Considering the massive number of potential devices and sporadic activity, the activity detection problem is essentially a compressive sensing problem.\par

If there is only one base station, by exploiting the sparse structure, sparsity constrained optimization methods have been developed for active user detection. Examples include orthogonal matching pursuit (OMP)\cite{schepker2013exploiting}, basis pursuit denoising (BPDN)\cite{wunder2014compressive,wunder2015compressive}, orthogonal least-squares (OLS)\cite{schepker2012compressive}, least absolute shrinkage and selection operator (LASSO)\cite{applebaum2012asynchronous,li2019activity}, and covariance-based\cite{8437359,9691883,liyangtransformer} methods. These methods only differ in how the sparsity constraint is transformed into an equivalent penalty term (e.g., $\ell_1$-norm in\cite{wunder2014compressive,wunder2015sparse,wunder2015compressive,applebaum2012asynchronous,8437359} or $\ell_{0}$-norm in\cite{schepker2012compressive,li2019activity,lin2023}).

\IEEEpubidadjcol
Sparse Bayesian learning (SBL) methods have also been used for activity detection, where statistical distribution of the sparse activity is incorporated in the probabilistic model. For example, Bernoulli-Gaussian distribution is frequently used for modeling sparse signals with a known activity probability. Based on this model, maximum a posteriori (MAP)\cite{zhu2010exploiting}, approximate message passing (AMP)\cite{liu2018massive,liu2018massive3,hannak2015joint,chen2017massive,chen2018sparse,ding2019sparsity} and expectation propagation (EP)-based algorithms\cite{ahn2019ep} have been proposed to detect user activity status. Besides, sparsity-promoting priors such as Gaussian-gamma distribution, which do not need the information of activity probability, are utilized for activity detection\cite{zhang2021joint,zhang2021low,wang2022double}, and variational inference is used for deriving the detection algorithms.

\begin{figure}[!t]
	\centering
	\includegraphics[width=3.5in]{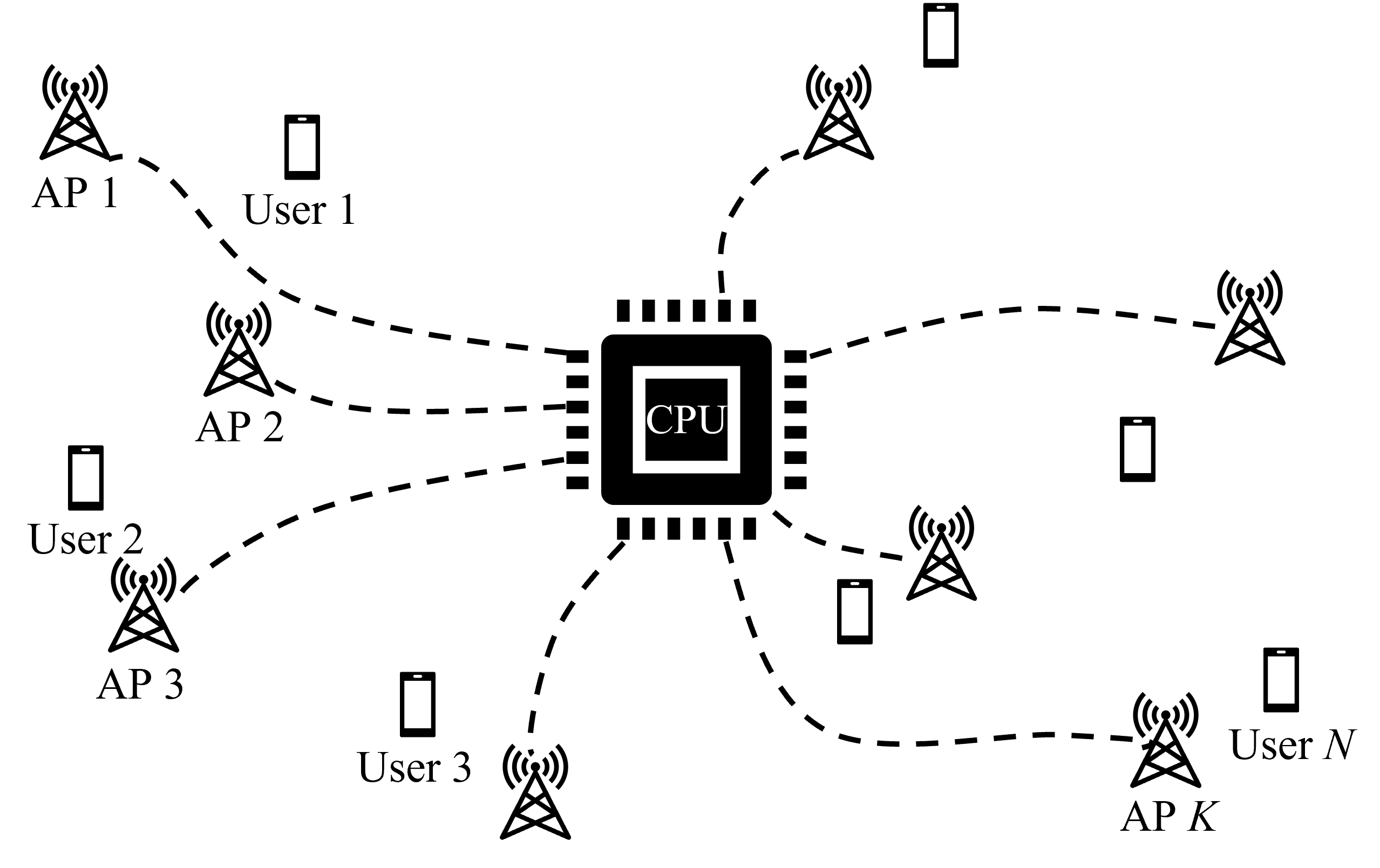}
	\caption{Cell-free network model.}
	\label{cellfreearc}
\end{figure}

While the above works set the foundation for activity detection in grant-free access, they all focus on a single cell setting. However, in reality, there 
are usually many access points (APs) in a modern wireless system. When each user communicates with its intended AP, it inevitably introduces inter-cell interference\cite{interdonato2019ubiquitous}. Recently, cell-free massive MIMO system was proposed to eliminate the cell boundaries, wherein all the APs cooperatively serve all the users in a wide area as shown in Fig. \ref{cellfreearc}. By connecting all APs to one central computing unit (CPU) such as a cloud-RAN (C-RAN) computer for handling the received signals jointly, cell-free networks provide better interference management and achieve higher throughput and scalability than single-cell systems or multi-cell systems where each AP serves its own group of users\cite{ngo2017cell,interdonato2019ubiquitous,zhou2018,zhou2014,tan2021}.

To apply existing activity detection methods to the cell-free setting, one idea is to estimate the activity status of each user independently at each AP, and then the final detection result is a weighted combination of the results from different APs with their corresponding reliabilities. This has in fact appeared in recent extensions of single-cell covariance-based algorithm\cite{xu2015active,liyangdcov}, variational inference algorithm with Gaussian-gamma prior\cite{crangg}, and AMP algorithm\cite{guo2021joint,guo2020sparse,guo2019distributed}. For more rigorous information combining, AMP-based likelihood ratio fusion has also been proposed\cite{multicell,dAMP,cranmp}. While these extensions provide workable solutions, they all ignore the fact that each user should only have a single activity status, thus leaving a tremendous potential for performance improvement. To the best of our knowledge, there is only one recent work addressing the activity status consistency, and it proposes a network-wide joint cost optimization based on the covariance-based method \cite{ganesan2021clustering}.

No matter whether activity status consistency is imposed, directly extending the covariance-based and AMP-based methods would inherit the dependence on many system parameters. For example, in the single-cell covariance-based method with a fusion step \cite{xu2015active}, noise variances at the APs, and the small-scale fading being Rayleigh fading are required. For the cell-free covariance-based method\cite{ganesan2021clustering}, additional knowledge of large-scale fading coefficients is also required. On the other hand, although the AMP methods \cite{guo2021joint,guo2020sparse,guo2019distributed,multicell,dAMP} would estimate the channel so that they are not highly dependent on the Rayleigh fading assumption, their formulation requires the knowledge of large-scale fading coefficients, noise variances at APs, and user activity probability.
	
Unfortunately, in practice, these system parameters are hard to obtain. For example, large-scale fading is not only determined by distance but also shadowing, which is random. Even though large-scale fading can be estimated\cite{lsfc}, estimation error is inevitable, with relative error ranging from 20\% up to 60\%\cite{9374476}. The situation is more challenging in cell-free networks as there are many more large-scale fading coefficients to be estimated, which may take a long time to accomplish. Furthermore, in grant-free network, the users could be Internet-of-Things (IoT) devices. Due to limited power of these devices, line-of-sight (LoS) channel is preferred. Therefore, it is expected that some channels might be Rayleigh fading while some others are Rician fading, and we even do not know which channel is Rayleigh fading and which one is not. This makes Rayleigh fading assumption no longer appropriate, and the covariance matrix assumed in the covariance-based method would be different from that of the received signal, which leads to severe degradation in the detection results.\par 
 
To reduce the reliance of activity detection algorithms on these system parameters, there have been intermittent recent efforts. For example, the LoS propagation has recently been taken into consideration in covariance-based methods\cite{Rician1,Rician2}. However, information about which users experiencing Rician fading and the respective Rician factors are still required, which regrettably is not easily achieved in practice. In addition, the AMP algorithm has recently been extended to estimate the activity probability, large-scale fading coefficients, and noise variance\cite{gmmvamp}. However, this extension only applies to single-cell settings.

In order to enforce activity status consistency in cell-free system and decouple from the knowledge of the large-scale fading coefficients, small-scale fading statistics, noise variance, and activity probability, we propose to jointly estimate user activity status together with all these unknown parameters, with specific regularizations on these parameters imposed from a Bayesian perspective. While probabilistic models have been proposed in previous works\cite{zhang2021joint,zhang2021low,wang2022double,crangg} for activity detection, they cannot be extended to the cell-free setting directly because the sparse activity of a particular user in cell-free system is coupled among all APs. Besides, due to the rigidity of the Gaussian-gamma prior employed in \cite{zhang2021joint,zhang2021low,wang2022double,crangg}, they are effective only if the user activity level is around 10\% to 15\%.
	
To overcome the limitations of existing Bayesian models, this paper proposes to enforce the coupled sparsity structures by introducing hierarchical priors with a common latent variable to the combined activity status and large-scale fadings from a user to all APs. Moreover, to enable the algorithm adapt to different user activity levels, we introduce the advanced generalized hyperbolic (GH) prior distribution, which is a highly adaptable sparsity-enhancing distribution\cite{thabane2004matrix,cheng2022towards}. With these two novelties, the activity status of each user is detected jointly by all APs, rather than each AP independently, and the resultant algorithm would maintain a high activity detection accuracy in a wide range of user activity probability. Furthermore, the proposed Bayesian model also includes noise variance and small-scale fading as random variables, with appropriate prior distributions imposed on them, such that the model would not rely on their precise knowledge.

Nevertheless, such sophisticated probabilistic model does not admit exact Bayesian inference, as the posterior distribution of the unknown variables cannot be obtained in closed-from. To get around this difficulty, we propose two algorithms. The first one is a block coordinate descent (BCD)-based MAP estimator, which updates each variable sequentially with the other variables fixed at the values given by the latest iteration. While the MAP estimator is simple in derivation and facilitates comparison with optimization-based algorithms, it can only provide point estimates to different parameters in each update step, and cannot fully utilize the information from the Bayesian model. Therefore, we further derive a second algorithm based on the variational inference technique, which seeks an approximate distribution that is close to the posterior distribution in Kullback-Leibler (KL) divergence sense. The obtained variational inference algorithm not only has a closed-form expression in each variable update, but also takes into account the uncertainties of other estimated parameters, which leads to more accurate detection performance than the MAP solution.

Simulation results show that the proposed methods outperform state-of-the-art Bayesian and non-Bayesian methods in terms of activity detection accuracy. Especially under imprecise knowledge of large-scale fading coefficients, small-scale fading statistics, noise variance, and user activity probability, the performance of the state-of-the-art covariance-based and AMP-based methods degrade significantly. In contrast, the proposed methods provide robust performance and achieve the lowest detection error without any knowledge of these parameters about the wireless environment.

Compared to the preliminary version\cite{icassp2024}, this paper includes detailed explanations of the probabilistic model, in particular on how to ensure the detected results of an inactive user would be consistent among all the APs while allowing significant variations of large-scale fadings of an active user to multiple APs, and on how the proposed model could provide adaption to different activity levels. Furthermore, this paper derives the variational inference algorithm which is an enhancement to the MAP algorithm presented in \cite{icassp2024}, provides more insights and properties of the proposed algorithms, and presents more comprehensive comparison with existing works.

The rest of this paper is organized as follows. Section \ref{sec:system model} describes the system model and two different likelihood functions. Section \ref{sec:modelandalgorithm} introduces the proposed probabilistic modeling and derives the MAP algorithm. Section \ref{sec:vi} presents the variational inference algorithm. Section \ref{sec:simulation} provides the simulation results and discussions. Conclusions are drawn in section \ref{sec:conclusion}.

\textit{Notations}: Bold-face upper-case letters denote matrices, bold-face lower-case letters denote vectors, and lower-case letters denote scalars. $\Bar{(\cdot)}$, $(\cdot)^{-1}$, $(\cdot)^{T}$, $(\cdot)^{H}$, $\text{Tr}(\cdot)$ denote conjugate, inverse, transpose, Hermitian transpose and trace operator respectively. $\mathbf{I}_M$ denotes the identity matrix with size $M$, and $\bm{0}$ denotes the all-zero vector. $\mathbb{E}\left[\cdot\right]$ or $\langle\cdot\rangle$ denotes expectation of random variables. $\|\cdot\|_0$ denotes $l_0$ norm, $\|\cdot\|_2$ denotes $l_2$ norm, and $\|\cdot\|_F$ denotes the Frobenius norm.
\section{System Model and Two Maximum Likelihood Estimators}
\label{sec:system model}
Consider a grant-free multiple access system in Fig. \ref{cellfreearc} which contains $N$ potential single-antenna users and $K$ APs each equipped with $M$ antennas. The channel between the $n$-th user and the $k$-th AP is modeled as $\sqrt{\beta_{kn}}\mathbf{g}_{kn}$, where $\beta_{kn}\in\mathbb{R}$ represents the large-scale fading, $\mathbf{g}_{kn}\in\mathbb{C}^M$ is the small-scale fading, and is independent with respect to $n$. We adopt a block-fading model on the channels, where $\left\{\sqrt{\beta_{kn}}\mathbf{g}_{kn}\right\}_{k=1,n=1}^{K,N}$ are constant in each coherence block.\par

Each user $n$ is assumed to transmit data sporadically and independently with an unknown probability. We define $a_n\in\left\{0,1\right\}$ as the activity indicator of user $n$ within a single coherence block, where $a_n=1$ means that the $n$-th user is active and $a_n=0$ otherwise. If the $n$-th user is active, it sends a unique pilot sequence $\mathbf{s}_n$ of length $L$. Assuming that the pilot sequences $\left\{\mathbf{s}_n\right\}_{n=1}^N$ from all $N$ devices are known, the received signal at the $k$-th AP $\mathbf{Y}_k\in\mathbb{C}^{L\times M}$ is modeled as
\begin{align}
	\label{receivedY}
	\mathbf{Y}_k &= \sum_{n=1}^Na_n\sqrt{\beta_{kn}}\mathbf{s}_n\mathbf{g}_{kn}^T+\mathbf{W}_k,
\end{align}
where $\mathbf{W}_k$ denotes the additive white Gaussian noise.

Consider the following assumptions:\\
\textbf{A1)} All the large-scale fading coefficients $\left\{\beta_{kn}\right\}_{k=1,n=1}^{K,N}$ between all the users and APs are precisely known;\\
\textbf{A2)} All the small-scale fading is Rayleigh fading with identity matrix as the covariance, i.e., $\mathbf{g}_{kn}\sim\mathcal{CN}(\mathbf{0},\mathbf{I}_M)$;\\
\textbf{A3)} The noise variance $\sigma^2$ is known.

Under the assumptions \textbf{A1} to \textbf{A3}, the covariance matrix of $\mathbf{Y}_k$ at the $k$-th AP can be shown to be\cite{ganesan2021clustering}
\begin{align}
	\label{covariancematrix}
	\mathbf{Q}_k=\sum_{n=1}^{N}a_{n}\beta_{kn}\mathbf{s}_n\mathbf{s}_n^H+\sigma^2\mathbf{I}_L,
\end{align}
and the likelihood function of the received signals from all APs $\mathcal{Y}=\left\{\mathbf{Y}_k\right\}_{k=1}^K$ is
\begin{align}
	\label{modifiedlikelihood}
	p(\mathcal{Y}|\left\{a_n\right\}_{n=1}^N)&=\prod_{k=1}^{K}\frac{1}{\left|\pi\mathbf{Q}_k\right|^M}\exp\left(-\text{Tr}\left(\mathbf{Q}_k^{-1}\mathbf{Y}_k\mathbf{Y}_k^H\right)\right).
\end{align}
Based on the covariance matrix (\ref{covariancematrix}), \cite{ganesan2021clustering} proposes to optimize the logarithm of (\ref{modifiedlikelihood}) with respect to the activity status $\left\{a_n\right\}_{n=1}^N$, and this is known as the covariance-based method:
\begin{align}
	\label{covarianceobj}
	\min_{\left\{a_n\right\}_{n=1}^N}&\sum_{k=1}^{K}\left[\log\left|\sum_{n=1}^{N}a_{n}\beta_{kn}\mathbf{s}_n\mathbf{s}_n^H+\sigma^2\mathbf{I}_L\right|\right.\nonumber\\
	&\left.+\text{Tr}\left(\left(\sum_{n=1}^{N}a_{n}\beta_{kn}\mathbf{s}_n\mathbf{s}_n^H+\sigma^2\mathbf{I}_L\right)^{-1}\dfrac{\mathbf{Y}_k\mathbf{Y}_k^H}{M}\right)\right]\nonumber\\
	\text{s.t. }&a_n\geq0\text{ for }n=1,2,\cdots,N.
\end{align}
From \eqref{covarianceobj}, it is clear that the covariance-based method heavily relies on the accuracy of the assumptions \textbf{A1}-\textbf{A3}. However, they may not hold in practice due to the inaccurate estimation of large-scale fading\cite{lsfc}, noise variance, and the presence of LoS path.\par
Notice that the violation of \textbf{A1} does not affect the covariance-based method much if it neglects the consistent activity status and each AP performs detection independently. In that case, there is no $k$ and thus $\beta_{kn}$ becomes $\beta_n$ that can be merged with $a_n$ to become a single variable for estimation\cite{9374476,covchen2022}. With an additional step of fusion, user activity detection in cell-free networks can be achieved without relying on the assumption \textbf{A1}. But for cell-free case as shown in \eqref{covarianceobj}, merging of $a_n$ with each $\beta_{kn}$ does not alleviate the lack of information on $\beta_{kn}$ so that assumption \textbf{A1} is indispensable for covariance-based method under cell-free systems if we impose activity status consistency for each user.

To decouple from \textbf{A1}-\textbf{A3}, we can write another likelihood function which explicitly depends on the activity status $\left\{a_n\right\}_{n=1}^N$, large-scale fading coefficients $\left\{\beta_{kn}\right\}_{k=1,n=1}^{K,N}$, small-scale fading $\left\{\mathbf{g}_{kn}\right\}_{k=1,n=1}^{K,N}$, and noise variance $\sigma^2$:
\begin{align}
	\label{likelihood}
	&p(\mathcal{Y}|\bm{\Theta})\nonumber\\
	&=\prod_k^K\prod_i^L\prod_j^M\mathcal{CN}\left(\left(\mathbf{Y}_k\right)_{ij}|\left(\sum_{n=1}^Na_n\sqrt{\beta_{kn}}\mathbf{s}_n\mathbf{g}_{kn}^T\right)_{ij},\sigma^2\right)\nonumber\\
	&=\dfrac{\exp\left(-\dfrac{1}{\sigma^2}\sum_{k=1}^K\left\|\mathbf{Y}_k-\sum_{n=1}^Na_n\sqrt{\beta_{kn}}\mathbf{s}_n\mathbf{g}_{kn}^T\right\|_F^2\right)}{\left(\sigma^2\pi\right)^{KLM}},
\end{align}
where $\bm{\Theta}=\left\{\left\{a_n\right\}_{n=1}^N,\left\{\beta_{kn}\right\}_{k=1,n=1}^{K,N},\left\{\mathbf{g}_{kn}\right\}_{k=1,n=1}^{K,N},\sigma^2\right\}$.

The maximum likelihood estimation of $\bm{\Theta}$ is achieved by maximizing \eqref{likelihood}. However, without any constraint, this optimization problem has a large solution space and it is challenging to get a solution. In order to shrink the solution space, specific regularizations on the problem are required. For example, utilizing the sparse structure of $a_n\sqrt{\beta_{kn}}\mathbf{g}_{kn}$, $\ell_0$-norm regularization can be introduced to improve the detection performance\cite{schepker2012compressive,li2019activity,lin2023}. In fact, the covariance-based method \eqref{covarianceobj} can be viewed as an extreme form of regularization on $\left\{\beta_{kn}\right\}_{k=1,n=1}^{K,N}$, $\left\{\mathbf{g}_{kn}\right\}_{k=1,n=1}^{K,N}$ and $\sigma^2$ by imposing assumptions \textbf{A1}-\textbf{A3}. As these assumptions may not strictly hold, one may wonder if we can impose regularizations on these parameters while still allowing uncertainty in our prior knowledge. We will employ probabilistic model to achieve this purpose in the following section.
\section{Encoding Prior Knowledge with Probabilistic Modeling}
\label{sec:modelandalgorithm}
Bayesian probabilistic models are learning models that assume the unknown variables obeying certain posterior distribution, which is composed of prior distributions reflecting our belief on the variables before any observation, and the likelihood function \eqref{likelihood} describing the observations. Consequently, the prior distributions provide a natural way of determining the form of regularizations on the unknown parameters.

Notice that the AMP-based method also comes from a Bayesian model which makes use of the Bernoulli-Gaussian prior distribution for the combined activity status and small-scale fading:
\begin{align}
	\label{bernoulligaussian}
	p\left(\mathbf{g}_{kn}\right)=\left(1-\epsilon_{kn}\right)\delta\left(\mathbf{g}_{kn}\right)+\epsilon_{kn}\mathcal{CN}\left(\mathbf{0},\mathbf{I}_M\right),
\end{align}
where $\delta(\cdot)$ is the Dirac delta function, and $\epsilon_{kn}$ is the activity probability of the $n$-th user at the $k$-th AP. By combining \eqref{likelihood} and \eqref{bernoulligaussian}, AMP algorithm seeks for the approximate posterior distribution that minimizes the mean square error\cite{liu2018massive}:
\begin{align}
	\label{MMSEproblem}
	\mathbb{E}_{\mathbf{g}_{kn}}\left[\left\|\hat{\mathbf{g}}_{kn}\left(\mathbf{Y}_k\right)-\mathbf{g}_{kn}\right\|_2^2\right].
\end{align}However, the Bernoulli-Gaussian model \eqref{bernoulligaussian} requires the knowledge of the user activity probability:
\begin{spacing}{1.5}
\noindent\textbf{A4)} User activity probability is known.
\end{spacing}
With the user activity probability may change over time, the assumption \textbf{A4} is not necessarily accurate. Although a recent algorithm GMMV-AMP\cite{gmmvamp} has been proposed to learn the activity probability, large-scale fading, and noise variance, it only applies to single cell networks. Even if we could combine the detection results from various APs, the consistent activity status among all APs is not fully utilized.\par
In the following, we will employ a prior distribution that not only ensures consistent activity status among all the APs, but also is flexible to cope with various degrees of user activity without the user activity probability specification. As a preview, the required input of different methods and whether each method utilizes consistent activity status are summarized in Table \ref{tab_algin}.
\begin{table*}[!t]
	\caption{Comparison of Algorithm Characteristics\label{tab_algin}}
	\centering
	\begin{tabular}{|c||c|c|c|c|c|c|c|c|}
		\hline
		Algorithm & $\mathcal{Y}$ & $\mathbf{S}$ & $\varepsilon$ & $\beta_{kn}$ & \thead{$\tau$} & \thead{Rayleigh \\ fading}& \thead{Activity status \\ consistency}\\
		\hline
		GHVI (proposed) & \checkmark & \checkmark & & & & &\checkmark\\
		MAP (proposed) & \checkmark & \checkmark & & & & &\checkmark\\
		Cell-free covariance-based method\cite{ganesan2021clustering} & \checkmark & \checkmark & & \checkmark & \checkmark & \checkmark & \checkmark\\
		\makecell{Single-cell covariance-based method plus decision fusion\cite{xu2015active,covchen2022}} & \checkmark & \checkmark & & & \checkmark & \checkmark &\\
		\makecell{Single-cell AMP plus LLR fusion\cite{dAMP,cranmp}} & \checkmark &\checkmark& \checkmark & \checkmark & \checkmark &  &\\
		\makecell{Extension of single-cell GMMV-AMP\cite{gmmvamp} with LLR fusion} & \checkmark &\checkmark& &  &  &  &\\
		\makecell{Single-cell Gaussian-gamma VI plus decision fusion\cite{crangg}} & \checkmark & \checkmark & & & &  &\\
		\hline
	\end{tabular}
\end{table*}
\subsection{Sparsity-enhancing Prior for Combined Activity Status and Large-scale Fading}
If assumption \textbf{A1} is violated, $\left\{\beta_{kn}\right\}_{k=1,n=1}^{K,N}$ are unknowns. To exploit the fact that $\left\{\beta_{kn}\right\}_{k=1}^K$ do not need to be estimated if user $n$ is inactive, we combine the activity status $a_n$ and large-scale fading coefficients $\beta_{kn}$ via $\gamma_{kn}=a_n\sqrt{\beta_{kn}}$, and this new variable follows the same sparsity pattern of $a_n$. We assume that \textbf{A4} is not satisfied, so Bernoulli model cannot be applied. Instead, we adopt a flexible sparsity inducing GH prior\cite{cheng2022towards} for $\left\{\gamma_{kn}\right\}_{k=1,n=1}^{K,N}$, which is a hierarchical construction: 
\begin{align}
	p(\gamma_{kn}|z_n)&=\mathcal{N}\left(\gamma_{kn}|0,z_n\right)\nonumber\\
	&=\dfrac{1}{\sqrt{2\pi z_n}}\exp\left(-\dfrac{\gamma_{kn}^2}{2z_n}\right),\label{gammaknprior}\\
	p(z_n)&=\text{GIG}\left(z_n|\eta_{n}^0,\psi_n^0,\lambda_n^0\right)\label{zprior}.
\end{align}
In \eqref{gammaknprior}, $\left\{z_n\right\}_{n=1}^N$ are latent variables representing the variance of the Gaussian distributions in this hierarchical construction. Applying a Gaussian prior on $\gamma_{kn}$ is equivalent to an $\ell_2$ regularization on $\gamma_{kn}$ with regularization parameter $z_n^{-1}/2$\cite{bishop2006pattern}. However, in contrast to conventional regularization, $z_n$ here obeys another distribution
\begin{align}
	\label{gig}
	\text{GIG}(z_n|\eta_n^0,\psi_n^0,\lambda_n^0)=&\dfrac{(\frac{\eta_n^0}{\psi_n^0})^{\frac{\lambda_n^0}{2}}}{2K_{\lambda_n^0}(\sqrt{\eta_n^0\psi_n^0})}z_n^{\lambda_n^0-1}\nonumber\\
	&\times\exp\left(-\frac{1}{2}(\eta_n^0z_n+\psi_n^0z_n^{-1})\right),
\end{align}
where $K(\cdot)$ is the modified Bessel function of the second kind, and $\left\{\eta_n^0, \phi_n^0, \lambda_n^0\right\}_{n=1}^N$ are hyper-parameters. In this way, the regularization parameter $z_n^{-1}/2$ will be automatically learned instead of being set at a fixed number. 

It is worth noticing in \eqref{gammaknprior} that $\left\{\gamma_{kn}\right\}_{k=1}^K$ are controlled by the same variance $z_n$. Therefore, a small $z_n$ will drive all the associated large-scale fading coefficients $\left\{\gamma_{kn}\right\}_{k=1}^K$ to be zero. This corresponds to the case that user $n$ is inactive. In contrast, if $z_n$ is large, the regularization effect is small and the posterior distribution of $\left\{\gamma_{kn}\right\}_{k=1}^K$ will be heavily influenced by the likelihood function \eqref{likelihood}, which will make the posterior mean and covariance of $\gamma_{kn}$ very different for different $k$. This is the way how $z_n$ enforces the consistency of the activity status of each user at different APs (i.e., when a user is inactive, the corresponding $\gamma_{kn}$ will be zero for all $k$), while allowing the posterior mean and covariance of $\gamma_{kn}$ to take different values for different $k$ to account for the significant differences of large-scale fading of a particular active user $n$ to different APs. This novel way of activity modeling in massive access has not been done before, since covariance-based method with consistent activity status for cell-free networks assumes $\left\{\beta_{kn}\right\}_{k=1,n=1}^{K,N}$ to be known.\par
Note that in the proposed model, $\gamma_{kn}$ and $\mathbf{g}_{kn}$ are multiplied together in the likelihood function \eqref{likelihood}. This could lead to an ambiguity in the product $\gamma_{kn}\mathbf{g}_{kn}$ in the sense that a $180\degree$ shift in the $\mathbf{g}_{kn}$ would end up in $\gamma_{kn}$, which makes it a negative value. This motives us to model $\gamma_{kn}$, which should be greater than or equal zero, using a two-sided distribution in (8). In case the learnt value of  $\gamma_{kn}$ is negative, the negative sign can be moved back to the $\mathbf{g}_{kn}$.\par
\begin{figure}[!t]
	\centering
	\includegraphics[width=3.5in]{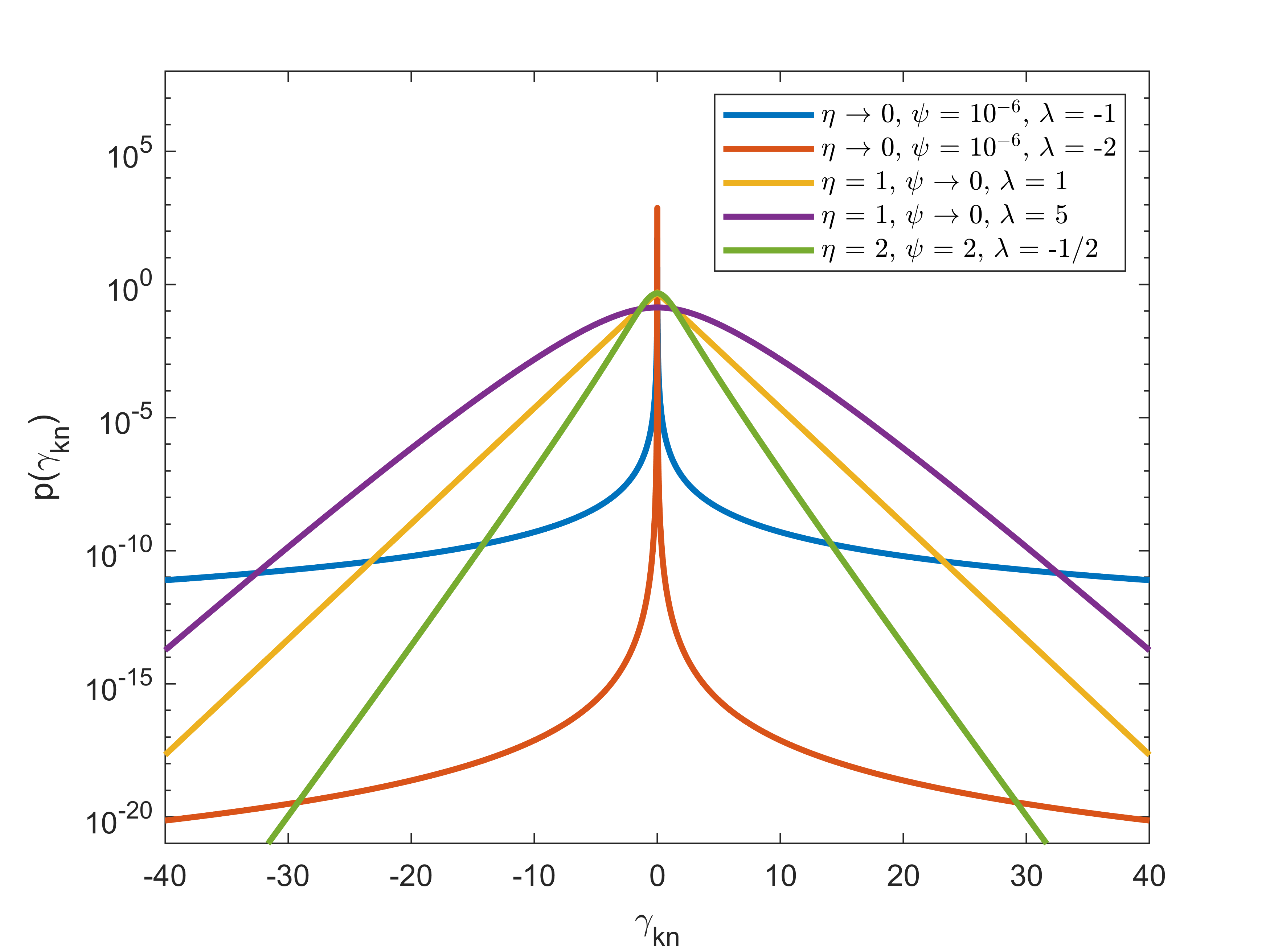}
	\caption{Marginal probability density function of GH prior with different values of hyper-parameters.}
	\label{figghprior}
\end{figure}

The generalized inverse Gaussian (GIG) distribution in \eqref{zprior} is a flexible hyper-prior for $z_n$. By using \eqref{gammaknprior}-\eqref{gig}, it can be shown that the marginal distribution of $\left\{\gamma_{kn}\right\}$ is\cite{cheng2022towards}
\begin{align}
	\label{ghprior}
	p\left(\gamma_{kn}\right)=&\int\mathcal{N}\left(\gamma_{kn}|z_n\right)\times\text{GIG}\left(z_n|\eta_n^0,\psi_n^0,\lambda_n^0\right)dz_n\nonumber\\
	=&\frac{\left(\eta_{n}^0\right)^{\frac{1}{4}}}{\sqrt{2\pi}}\frac{\left(\psi_n^0\right)^{-\frac{\lambda_n^0}{2}}}{K_{\lambda_n^0}\left(\sqrt{\eta_n^0\psi_n^0}\right)}\nonumber\\
	&\times\frac{K_{\lambda_n^0-\frac{1}{2}}\left(\sqrt{\eta_n^0\left(\psi_n^0+\left\|\gamma_{kn}\right\|_2^2\right)}\right)}{\psi_n^0+\left\|\gamma_{kn}\right\|_2^2}.
\end{align}
To show the flexibility of the GH prior, the marginal probability density function of $\gamma_{kn}$ \eqref{ghprior} is shown in Fig. \ref{figghprior} under different hyper-parameters, from which we can see that it exhibits a wide range of sparsity behavior. In fact, the hyper-parameters $\left\{\eta_n^0,\psi_n^0,\lambda_n^0\right\}$ correspond to the tail, peakness, and shape parameters, of which different choices lead to \eqref{ghprior} being reduced to different sparsity-enhancing priors, including but not limited to hyperbolic distribution, normal-inverse Gaussian distribution, and student-t distribution\cite{thabane2004matrix}. Especially when $\eta_n^0\rightarrow0$ and $\lambda_n^0<0$, the GH prior reduces to Gaussian-gamma prior used in\cite{zhang2021joint,zhang2021low,wang2022double,crangg}. This makes the hierarchical construction in \eqref{gammaknprior} and \eqref{zprior} highly flexible in the modeling\cite{cheng2022towards} so that it can be used to learn different sparsity levels of user activities.

On the other hand, previous research\cite{etaprior} shows that among the three hyper-parameters, $\eta_{n}^0$ has the largest effect to the shape of \eqref{ghprior}. In order to allow adaptable sparsity level to match with that of the received signals, we follow the related works\cite{cheng2022towards,beal2003variational} and introduce a Gamma distribution for $\eta_n^0$:
\begin{align}
	\label{anprior}
	p(\eta_n^0)=Ga\left(\eta_n^0|\kappa_{1},\kappa_{2}\right)=\dfrac{1}{\Gamma(\kappa_1)}{\eta_n^0}^{\kappa_{1}-1}\exp\left(-\kappa_{2}\eta_n^0\right)\kappa_{2}^{\kappa_{1}}.
\end{align}
In general, the setting of $\left\{\kappa_{1},\kappa_{2},\psi_n^0,\lambda_n^0\right\}$ should reflect our prior knowledge of $z_n$ before any observation of the received signals. As we have no prior knowledge on the activity probability, we should set them to be near-zero values (e.g., $10^{-6}$) so that \eqref{ghprior}-\eqref{anprior} represent non-informative priors. This also means that we let the hyper-parameters in the posterior distribution of $\gamma_{kn}$ primarily influenced by the observed data rather than by the prior\cite{bishop2006pattern}.
\subsection{Modeling of Small-scale Fading Prior}
In IoT setting, to reduce energy for data communication, LoS channel is preferred, making the channel amplitude obey Rician model, which is equivalent to the complex-valued channel being a non-zero mean complex Gaussian vector. However, before any observation, we do not know which user experiences Rician fading and the corresponding Rician factor. Without any prior knowledge about the LoS propagation, we place a complex Gaussian prior with zero mean and identity covariance on $\mathbf{g}_{kn}$:
\begin{align}
	\label{channelprior}
	p\left(\mathbf{g}_{kn}\right)=\mathcal{CN}\left(\mathbf{g}_{kn}|\mathbf{0},\mathbf{I}_{M}\right)=\dfrac{1}{\pi^M}\exp\left(-\left\|\mathbf{g}_{kn}\right\|_2^2\right).
\end{align}

While it seems that we are using Rayleigh fading in this prior distribution for the small-scale fading, an important point is that Bayesian framework does not directly perform inference using prior distribution. Instead, at the inference stage, information from the likelihood function would also contribute to the posterior distribution. If the specific channel is not of zero mean and the covariance is not of identity matrix, it would be reflected in the posterior distribution (see \eqref{updatemug} and \eqref{updatesigmag}).\par 
Therefore, although the prior distribution \eqref{channelprior} looks like we are building on Rayleigh fading assumption, at the inference stage, the learned channel statistics might not be Rayleigh fading anymore. Furthermore, as the channel statistics at the posterior distribution is dominantly influenced by the observations, we do not need to specify which channel has LoS propagation and the corresponding Rician factors in the prior distribution. This makes the assumption \textbf{A2} no longer necessary.
\subsection{Modeling Unknown Noise Precision}
For the noise variance $\sigma^2$, for notational convenience, we set $\tau$ as its inverse (known as precision), i.e., $\tau=\sigma^{-2}$, and model it as Gamma distributed:
\begin{align}
	\label{noiseprior}
	p(\tau)&=Ga\left(\tau|c,d\right)=\dfrac{1}{\Gamma(c)}\tau^{c-1}\exp\left(-d\tau\right)d^c,
\end{align}
where $\{c,d\}$ are the hyper-parameters of the Gamma distribution.\par 
This particular prior is chosen since it has non-negative support, which is consistent with the nature of noise precision. Furthermore, it is known that Gamma distribution is conjugate to the Gaussian likelihood function in \eqref{likelihood}, which facilitates the subsequent derivations of inference step for $\tau$. As we have no prior knowledge of the noise power before any observation, the hyper-parameters $\{c,d\}$ are set as near-zero values (e.g., $10^{-6}$) to express non-informative prior. In this way, the hyper-parameters of the posterior distribution would be primarily determined by the observations at the inference stage. The estimation of noise variance makes the assumption \textbf{A3} redundant.
\subsection{Posterior Distribution and the MAP Algorithm}
Combining the likelihood function of the received signals \eqref{likelihood} and the prior distributions of the variables \eqref{gammaknprior}-\eqref{zprior}, \eqref{anprior}-\eqref{noiseprior}, the joint distribution can be expressed as
\begin{align}
	\label{jointdistribution}
	p(\mathcal{Y},\bm{\Theta})=&\prod_{k=1}^{K}p\left(\mathbf{Y}_k|\left\{\gamma_{kn}\right\}_{n=1}^N,\left\{\mathbf{g}_{kn}\right\}_{n=1}^N,\tau\right)Ga\left(\tau|c,d\right)\nonumber\\
	&\times\prod_{k=1}^{K}\prod_{n=1}^{N}\mathcal{N}\left(\gamma_{kn}|0,z_n\right)\prod_{n=1}^{N}\text{GIG}\left(z_n|\eta_n^0,\psi_n^0,\lambda_n^0\right)\nonumber\\
	&\times \prod_{n=1}^{N}Ga\left(\eta_n^0|\kappa_{1},\kappa_{2}\right)\prod_{k=1}^{K}\prod_{n=1}^{N}\mathcal{CN}\left(\mathbf{g}_{kn}|\mathbf{0},\mathbf{I}_M\right).
\end{align}
The posterior distribution $p(\bm{\Theta}|\mathcal{Y})$ is then given by:
\begin{align}
	\label{posterior}
	p\left(\bm{\Theta}|\mathcal{Y}\right)=\dfrac{p\left(\mathcal{Y},\bm{\Theta}\right)}{\int p\left(\mathcal{Y},\bm{\Theta}^{\prime}\right)d\bm{\Theta}^{\prime}}.
\end{align}

In Bayesian model, the formal way to obtain inference on different variables is to marginalize the posterior distribution in \eqref{posterior} with respect to each variable. However, since the probabilistic model is complicated, the denominator of \eqref{posterior} does not have a closed-form expression. One way to get around this challenge is to use the MAP estimator, which is obtained by maximizing the posterior distribution \eqref{posterior} with respect to $\bm{\Theta}$. Due to the fact that the denominator of \eqref{posterior} is always positive and does not depend on $\bm{\Theta}$, the intractable integration in the denominator \eqref{posterior} is bypassed and the MAP estimator is equivalent to maximizing the joint distribution \eqref{jointdistribution}.\par 
Taking the logarithm of \eqref{jointdistribution}, the MAP optimization problem is 
\begin{align}
	\label{map}
	\max_{\bm{\Theta}}&\; KLM\ln\tau-\tau\sum_{k=1}^{K}\left\|\mathbf{Y}_k-\sum_{n=1}^{N}\gamma_{kn}\mathbf{s}_n\mathbf{g}_{kn}^T\right\|_F^2\nonumber\\
	&+\sum_{k=1}^{K}\sum_{n=1}^{N}\left(\frac{\ln z_n^{-1}-z_n^{-1}\gamma_{kn}^2}{2}\right)+\sum_{k=1}^{K}\sum_{n=1}^{N}\left(-\left\|\mathbf{g}_{kn}\right\|_2^2\right)\nonumber\\
	&+\sum_{n=1}^{N}\left\{\dfrac{\lambda_n^0}{2}\ln\dfrac{\eta_n^0}{\psi_n^0}-\ln\left[2K_{\lambda_n^0}\left(\sqrt{\eta_n^0\psi_n^0}\right)\right]+\left(\lambda_n^0-1\right)\right.\nonumber\\
	&\left.\times\ln z_n-\dfrac{1}{2}\left(\eta_n^0z_n+\psi_n^0z_n^{-1}\right)\right\}+\sum_{n=1}^{N}\left[\left(\kappa_{1}-1\right)\ln \eta_n^0\right.\nonumber\\
	&\left.-\kappa_{2}\eta_n^0\right]+\left(c-1\right)\ln\tau-d\tau+\text{const}.
\end{align}
The problem \eqref{map} can be solved via the BCD method in which each block coordinate (variable) is optimized sequentially with other variables fixed at their last updated values. In updating each block of variables, the objective function in \eqref{map} is differentiated and set to zero. Except $\left\{\eta_n\right\}_{n=1}^N$ are updated by the gradient descent algorithm, all other variables can be updated with closed-form expressions as follows.\\
\underline{\textbf{Update of} $\left\{\gamma_{kn}\right\}$:}
\begin{align}
	\label{mapupdategamma}
	\hat{\gamma}_{kn}=&\tau\text{Tr}\left\{\Re\left\{\Bar{\mathbf{g}}_{kn}\mathbf{s}_n^H\left[\mathbf{Y}_k-\sum_{m\neq n}\gamma_{km}\mathbf{s}_m\mathbf{g}_{km}^T\right]\right\}\right\}\nonumber\\
	&\times\left[\tau\left(\mathbf{s}_n^H\mathbf{s}_n\right)\left(\mathbf{g}_{kn}^H\mathbf{g}_{kn}\right)+\frac{z_n^{-1}}{2}\right]^{-1}.
\end{align}
\underline{\textbf{Update of} $\left\{\mathbf{g}_{kn}\right\}$:}
\begin{align}
	\label{mapupdateg}
	\hat{\mathbf{g}}_{kn}=&\tau\gamma_{kn}\left[\tau\gamma_{kn}^2\left(\mathbf{s}_n^H\mathbf{s}_n\right)+1\right]^{-1}\mathbf{I}_M\nonumber\\
	&\times\left(\mathbf{Y}_k^T-\sum_{m\neq n}\gamma_{km}\mathbf{g}_{km}\mathbf{s}_m^T\right)\bar{\mathbf{s}}_n.
\end{align}
\underline{\textbf{Update of} $\left\{z_{n}\right\}$:}
\begin{align}
	\label{mapupdatez}
	\hat{z}_n=&\dfrac{\hat{\lambda}_n+\sqrt{\hat{\lambda}_n^2+\eta_n^0\psi_n^0+\eta_n^0\sum_{k=1}^{K}\gamma_{kn}^2}}{\eta_n^0},
\end{align}
where $\hat{\lambda}_n=\lambda_n^0-\frac{K}{2}-1$.\\
\underline{\textbf{Update of} $\left\{\eta_{n}^0\right\}$:} Due to the lack of closed-form solution, $\hat{\eta}_n^0$ is updated via gradient descent until convergence:
\begin{align}
	&\hat{\eta}_n^0(t+1)=\eta_n^0(t)+\alpha\Delta\eta_n^0(t),\label{mapupdateeta}\\
	&\Delta\eta_n^0(t)=\dfrac{2\kappa_{1}-2+\lambda_n^0}{2\eta_{n}^0(t)}-\kappa_{2}-\frac{1}{2}z_n-\dfrac{1}{K_{\lambda_n^0}\left(\sqrt{\eta_n^0(t)\psi_n^0}\right)}\nonumber\\
	&\times\left(\dfrac{\lambda_n^0K_{\lambda_n^0}\left(\sqrt{\eta_n^0(t)\psi_n^0}\right)}{2\eta_n^0(t)}-\dfrac{\psi_n^0K_{\lambda_n^0+1}\left(\sqrt{\eta_n^0(t)\psi_n^0}\right)}{2\sqrt{\eta_n^0(t)\psi_n^0}}\right),\label{mapetagradient}
\end{align}
where $t$ denotes the gradient descent iteration number, and $\alpha$ is the step size.\\
\underline{\textbf{Update of} $\tau$:}
\begin{align}
	\label{mapupdatetau}
	\hat{\tau}=&\dfrac{KLM+c-1}{\sum_{k=1}^{K}\left\|\mathbf{Y}_k-\sum_{n=1}^{N}\gamma_{kn}\mathbf{s}_n\mathbf{g}_{kn}^T\right\|_F^2+d}.
\end{align}

After convergence, the activity status of user $n$ is determined by comparing $\hat{z}_n$ against a threshold $\rho$ (i.e., user $n$ is active if $\hat{z}_n>\rho$, otherwise it is inactive). The complete MAP algorithm is summarized in Algorithm \ref{mapalgorithm}. It resembles a conventional optimization-based algorithm, but with the form of regularization determined by the probabilistic model. Furthermore, the hyper-parameters of the MAP algorithm are automatically learned. This is not achievable from pure optimization perspective. Simulation results in Section \ref{sec:simulation} show that the MAP algorithm outperforms the state-of-the-art covariance-based and AMP-based methods (see Figs. \ref{figcf_PMDPFA}-\ref{figcf_PMDPFAAP}), demonstrating the effectiveness of the proposed probabilistic model.
\begin{algorithm}[!t]
	\SetAlgoLined
	\KwIn{The received signals $\left\{\mathbf{Y}_k\right\}_{k=1}^K$ and the pilot sequences $\left\{\mathbf{s}_n\right\}_{n=1}^N$.}
	
	\textbf{initialization:} $c=d=\kappa_{1}=\kappa_{2}=\eta_{n}^0=\psi_n^0=\lambda_n^0=10^{-6}$, $\left\{\gamma_{kn}=0\right\}_{k=1,n=1}^{K,N}$, $\left\{\mathbf{g}_{kn}\sim\mathcal{CN}\left(\mathbf{0},\mathbf{I}_M\right)\right\}_{k=1,n=1}^{K,N}$, $\left\{z_n=1\right\}_{n=1}^N$, $\tau=\frac{KLM}{\sum_{k=1}^{K}\left\|\mathbf{Y}_k\right\|_F^2}$;
	
	\Repeat{Convergence or reach maximum iterations}{
		Calculate $\left\{\gamma_{kn}\right\}_{k=1,n=1}^{K,N}$ via \eqref{mapupdategamma};
		
		Calculate $\left\{\mathbf{g}_{kn}\right\}_{k=1,n=1}^{K,N}$ via \eqref{mapupdateg};
		
		Calculate $\left\{z_n\right\}_{n=1}^N$ via \eqref{mapupdatez};
		
		\Repeat{Convergence or reach maximum iterations}{\For{$n=1,2,\dots,N$}{
				Calculate $\eta_{n}^0$ via \eqref{mapupdateeta} and \eqref{mapetagradient}
			}
		}
		
		Calculate $\tau$ via \eqref{mapupdatetau};
	}
	\KwOut{For $n=1,\dots,N$, activity indicator $a_n=1$ if $\hat{z}_n>\rho$, otherwise $a_n=0$.}
	\caption{BCD-based MAP algorithm}
	\label{mapalgorithm}
\end{algorithm}

Despite showing impressive performance, the MAP algorithm only provides point estimates and the variability of all the unknown parameters is dismissed in the learning procedure. However, a probabilistic model contains information about the uncertainties of various variables, which provide much richer information than point estimates. But as mentioned above, the intractable integration in the denominator of \eqref{posterior} prohibits us from taking the full Bayesian approach. In the next section, we will take a variational Bayesian approach to overcome this challenge. 
\section{Bayesian Variational Inference}
\label{sec:vi}
As an efficient method with theoretical guarantee of convergence, variational inference is typically used to find an approximation for the posterior distribution \eqref{posterior}. In variational inference, we seek a variational distribution $Q(\bm{\Theta})$ that minimizes the Kullback-Leibler (KL) divergence from the posterior distribution:
\begin{align}
	\label{viproblem}
	\min_{Q\left(\bm{\Theta}\right)}-\mathbb{E}_{Q\left(\bm{\Theta}\right)}\left\{\ln\dfrac{p\left(\bm{\Theta}|\mathcal{Y}\right)}{Q\left(\bm{\Theta}\right)}\right\}.
\end{align}

Without any restriction on $Q(\bm{\Theta})$, the $Q(\bm{\Theta})$ that minimizes \eqref{viproblem} is in fact $p(\bm{\Theta}|\mathcal{Y})$, which goes back to our original intractable problem. Therefore, $Q(\bm{\Theta})$ is usually optimized under the mean-field assumption\cite{bishop2006pattern}, which assumes
\begin{align}
	\label{meanvi}
	Q\left(\bm{\Theta}\right)&=\prod_{j=1}^{J}Q\left(\bm{\Theta}_j\right),
\end{align}
where $\bigcup\limits_{j=1}^J\bm{\Theta}_j=\bm{\Theta}$ and $\bigcap\limits_{j=1}^J\bm{\Theta}_j=\text{\O}$. The minimization of \eqref{viproblem} under the constraint \eqref{meanvi} can be solved iteratively block by block. For each block, the optimal solution is\cite{bishop2006pattern}
\begin{align}
	\label{meanvisolution}
	Q^*\left(\bm{\Theta}_j\right)=\dfrac{\exp\left(\mathbb{E}_{\prod_{i\neq j}Q\left(\bm{\Theta}_i\right)}\left[\ln p\left(\mathcal{Y},\bm{\Theta}\right)\right]\right)}{\int\exp\left(\mathbb{E}_{\prod_{i\neq j}Q\left(\bm{\Theta}_i^{\prime}\right)}\left[\ln p\left(\mathcal{Y},\bm{\Theta}^{\prime}\right)\right]\right)d\bm{\Theta}_j^{\prime}},
\end{align}
where $\mathbb{E}_{\prod_{i\neq j}Q(\bm{\Theta}_i)}[\cdot]$ denotes the expectation with respect to the variational distribution over all variables except $\bm{\Theta}_j$.

By substituting \eqref{jointdistribution} into \eqref{meanvisolution}, and using the mean-field
\begin{align}
	\label{meanviexpansion}
	Q\left(\bm{\Theta}\right)=&\prod_{k=1}^K\prod_{n=1}^NQ\left(\gamma_{kn}\right)\prod_{n=1}^NQ\left(z_n\right)\prod_{n=1}^{N}Q\left(\eta_n^0\right)\nonumber\\
	&\times \prod_{k=1}^K\prod_{n=1}^NQ\left(\mathbf{g}_{kn}\right)Q\left(\tau\right),
\end{align}
the closed-form update for each latent variable can be obtained, and are listed below. The detailed derivations are given in the Appendix.\\
\underline{\textbf{Update of} $\left\{Q(\gamma_{kn})\right\}$:} Each $\gamma_{kn}$ follows a Gaussian distribution, $Q(\gamma_{kn})=\mathcal{N}(\gamma_{kn}|\hat{\mu}_{kn}^{\gamma},\hat{\sigma}_{kn}^{\gamma})$ with $\hat{\mu}_{kn}^{\gamma}$ and $\hat{\sigma}_{kn}^{\gamma}$ respectively given by
\begin{align}
	\hat{\mu}_{kn}^{\gamma}=&2\text{Tr}\left\{\Re\left\{\langle\mathbf{g}_{kn}^T\rangle^H\mathbf{s}_n^H\left[\mathbf{Y}_k-\sum_{m\neq n}\langle\gamma_{km}\rangle\mathbf{s}_m\langle\mathbf{g}_{km}^T\rangle\right]\right\}\right\}\nonumber\\
	&\times\langle\tau\rangle\hat{\sigma}_{kn}^{\gamma},\label{updatemugamma}\\
	\hat{\sigma}_{kn}^{\gamma}=&\frac{1}{2}\left[\langle\tau\rangle(\mathbf{s}_n^H\mathbf{s}_n)\langle\mathbf{g}_{kn}^H\mathbf{g}_{kn}\rangle+\frac{\langle z_n^{-1}\rangle}{2}\right]^{-1}.\label{updatesigmagamma}
\end{align}
\underline{\textbf{Update of} $\left\{Q(\mathbf{g}_{kn})\right\}$:} Each $\mathbf{g}_{kn}$ follows a complex Gaussian distribution, $Q(\mathbf{g}_{kn})=\mathcal{CN}(\mathbf{g}_{kn}|\hat{\mu}_{kn}^{\mathbf{g}},\hat{\Sigma}_{kn}^{\mathbf{g}})$ with $\hat{\mu}_{kn}^{\mathbf{g}}$ and $\hat{\Sigma}_{kn}^{\mathbf{g}}$ given by
\begin{align}
	\hat{\mu}_{kn}^{\mathbf{g}}=&\langle\tau\rangle\langle\gamma_{kn}\rangle\hat{\Sigma}_{kn}^{\mathbf{g}}\left(\mathbf{Y}_k^T-\sum_{m\neq n}\langle\gamma_{km}\rangle\langle\mathbf{g}_{km}\rangle\mathbf{s}_m^T\right)\Bar{\mathbf{s}}_n,\label{updatemug}\\
	\hat{\Sigma}_{kn}^{\mathbf{g}}=&\left[\langle\tau\rangle\langle\gamma_{kn}^2\rangle(\mathbf{s}_n^H\mathbf{s}_n)+1\right]^{-1}\mathbf{I}_M.\label{updatesigmag}
\end{align}
\underline{\textbf{Update of} $\left\{Q(z_n)\right\}_{n=1}^N$:} Each $z_n$ follows a generalized inverse Gaussian distribution, $Q(z_n)=\text{GIG}(z_n|\hat{\eta}_n,\hat{\psi}_n,\hat{\lambda}_n)$ 
\begin{align}
	\hat{\eta}_n=&\langle \eta_n^0\rangle,\label{updatea}\\
	\hat{\psi}_n=&\psi_n^0+\sum_{k=1}^K\langle\gamma_{kn}^2\rangle,\label{updateb}\\
	\hat{\lambda}_n=&\lambda_n^0-\frac{K}{2}.\label{updatelambda}
\end{align}
\underline{\textbf{Update of} $\left\{Q(\eta_n^0)\right\}_{n=1}^N$:} Each $\eta_n^0$ follows a Gamma distribution, $Q(\eta_n^0)=Ga(\eta_n^0|\hat{\kappa}_{n1},\hat{\kappa}_{n2})$ where
\begin{align}
	\hat{\kappa}_{n1}=&\kappa_{1}+\frac{\lambda_n^0}{2},\label{updatek1}\\
	\hat{\kappa}_{n2}=&\kappa_{2}+\frac{\langle z_n\rangle}{2},\label{updatek2}
\end{align}
and the validity of $Q\left(\eta_{n}^0\right)$ is ensured when $\kappa_{1}>-\lambda_n^0/2$ and $\kappa_{2}>0$.\\
\underline{\textbf{Update of} $Q(\tau)$:} The noise precision $\tau$ follows a Gamma distribution, $Q(\tau)=Ga(\tau|\hat{c},\hat{d})$ where
\begin{align}
	\hat{c}=&LM+c,\label{updatec}\\
	\hat{d}=&\sum_{k}^{K}\sum_{n=1}^{N}\left\{\left(\mathbf{s}_n^H\mathbf{s}_n\right)\left[\hat{\sigma}_{kn}^{\gamma}\langle\mathbf{g}_{kn}^H\rangle\langle\mathbf{g}_{kn}\rangle+\langle\gamma_{kn}\rangle^2\text{Tr}\left(\hat{\Sigma}_{kn}^{\mathbf{g}}\right)\right.\right.\nonumber\\
	&\left.\left.+\hat{\sigma}_{kn}^{\gamma}\hat{\Sigma}_{kn}^{\mathbf{g}}\right]\right\}+\sum_{k=1}^{K}\left\|\mathbf{Y}_k-\sum_{n=1}^{N}\langle\gamma_{kn}\rangle\mathbf{s}_n\langle\mathbf{g}_{kn}^T\rangle\right\|_F^2+d.\label{updated}
\end{align}

In the update equations \eqref{updatemugamma}-\eqref{updated}, there are a number of expectations to be computed, and they are given by $\langle\gamma_{kn}\rangle=\hat{\mu}_{kn}^{\gamma}$, $\langle\mathbf{g}_{kn}\rangle=\hat{\mu}_{kn}^{\mathbf{g}}$, $\langle z_n^{-1}\rangle=\left(\frac{\hat{\psi}_n}{\hat{\eta}_n}\right)^{-\frac{1}{2}}\frac{K_{\hat{\lambda}_n+1}\left(\sqrt{\hat{\eta}_n\hat{\psi}_n}\right)}{K_{\hat{\lambda}_n}\left(\sqrt{\hat{\eta}_n\hat{\psi}_n}\right)}$, $\langle z_n\rangle=\left(\frac{\hat{\psi}_n}{\hat{\eta}_n}\right)^{\frac{1}{2}}\frac{K_{\hat{\lambda}_n+1}\left(\sqrt{\hat{\eta}_n\hat{\psi}_n}\right)}{K_{\hat{\lambda}_n}\left(\sqrt{\hat{\eta}_n\hat{\psi}_n}\right)}$, $\langle \eta_n^0\rangle=\hat{\kappa}_{n1}/\hat{\kappa}_{n2}$, and $\langle\tau\rangle=\hat{c}/\hat{d}$.\par
\begin{algorithm}[!t]
	\SetAlgoLined
	\KwIn{The received signals $\left\{\mathbf{Y}_k\right\}_{k=1}^K$ and the pilot sequences $\left\{\mathbf{s}_n\right\}_{n=1}^N$.}
	
	\textbf{initialization:} $c=d=\kappa_{1}=\kappa_{2}=\eta_{n}^0=\psi_n^0=\lambda_n^0=10^{-6}$, $\left\{\gamma_{kn}=0\right\}_{k=1,n=1}^{K,N}$, $\left\{\mathbf{g}_{kn}\sim\mathcal{CN}\left(\mathbf{0},\mathbf{I}_M\right)\right\}_{k=1,n=1}^{K,N}$, $\left\{z_n=1\right\}_{n=1}^N$, $\tau=\frac{KLM}{\sum_{k=1}^{K}\left\|\mathbf{Y}_k\right\|_F^2}$;
	
	\Repeat{Convergence or reach maximum iterations}{
		Update $\left\{Q(\gamma_{kn})\right\}_{k=1,n=1}^{K,N}$ via \eqref{updatemugamma} and \eqref{updatesigmagamma};
		
		Update $\left\{Q(\mathbf{g}_{kn})\right\}_{k=1,n=1}^{K,N}$ via \eqref{updatemug} and \eqref{updatesigmag};
		
		Update $\left\{Q(z_n)\right\}_{n=1}^{N}$ via \eqref{updatea}-\eqref{updatelambda};
		
		Update $\left\{Q(\eta_n^0)\right\}_{n=1}^{N}$ via \eqref{updatek1} and \eqref{updatek2};
		
		Update $Q(\tau)$ via \eqref{updatec} and \eqref{updated};
	}
	\KwOut{For $n=1,\dots,N$, activity indicator $a_n=1$ if $\langle z_n\rangle>\rho$, otherwise $a_n=0$.}
	\caption{GHVI algorithm}
	\label{vialgorithm}
\end{algorithm}

After convergence, the activity status of user $n$ is determined by comparing $\langle z_n\rangle$ against a threshold $\rho$ (i.e., user $n$ is active if $\langle z_n\rangle>\rho$, otherwise it is inactive). The complete algorithm (we term it as GHVI) is summarized in Algorithm \ref{vialgorithm}. Various properties of GHVI are presented below.
\subsubsection{Convergence Property}
When other variational distributions of $\left\{\bm{\Theta}_i\right\}_{i\neq j}$ are fixed, the problem \eqref{viproblem} is convex with respect to $Q(\bm{\Theta}_j)$\cite{boyd2004convex}. In one iteration, the KL divergence is optimized with respect to $Q(\bm{\Theta}_j)$ one block at a time using \eqref{meanvisolution}. Since this update is under the BCD framework, the convergence of the proposed method is guaranteed\cite{bishop2006pattern}. In implementation, the algorithm stops when the relative change in the recovered signals is smaller than $10^{-4}$.
\subsubsection{Sparsity and Large-scale Fading Learning}
During the inference procedure, the means of the latent variables $\left\{z_n\right\}_{n=1}^N$ and their inverses are learned from the observations. As GH prior is sparsity-enhancing, some of the $\left\{\langle z_n^{-1}\rangle\right\}_{n=1}^N$ will be updated to large numbers\cite{cheng2022towards}, e.g., in the order of $10^2$. As the variance of $\left\{\gamma_{kn}\right\}_{k=1}^K$ in \eqref{updatesigmagamma} is determined by $\langle z_n^{-1}\rangle$, a large $\langle z_n^{-1}\rangle$ will shrink all the variances of $\left\{\gamma_{kn}\right\}_{k=1}^K$ and at the same time drive the means $\left\{\hat{\mu}_{kn}^{\gamma}\right\}_{k=1}^K$ in \eqref{updatemugamma} to zero. On the other hand, the rest of $\left\{\langle z_n^{-1}\rangle\right\}_{n=1}^N$ would be learned to be small values to allow large variation of $\left\{\gamma_{kn}\right\}_{k=1}^K$ for the user $n$. As a result, the sparsity structure of device activities is revealed in the learned $\left\{z_n\right\}_{n=1}^{N}$.
\subsubsection{Hyper-parameter Settings in Priors}
The prior distributions we place on the unknown variables reflect our beliefs about the variables before any observations. In this paper, we assume that we have no knowledge on the activity probabilities and noise variance. Therefore, in the priors \eqref{gig}, \eqref{anprior}, and \eqref{noiseprior}, the hyper-parameters are set as near-zero values (e.g., $10^{-6}$) to indicate non-informative priors. Notice that this does not mean the non-informative prior has no purpose and can be removed. This only means that the prior distributions have little effect on the posterior distributions or the variational distributions. At the inference stage, the information from observations, which is represented in the likelihood function, is combined with the prior distributions via Bayes’ theorem to form the posterior distribution from which it becomes informative. In this way, the updated hyper-parameters are dominantly determined from the observed data via \eqref{updatea}-\eqref{updated}.
\subsubsection{Uncertainty Utilization in GHVI}
In each iteration of GHVI (i.e., equations \eqref{updatemugamma}-\eqref{updated}), the variational distribution that approximates the posterior distribution of each variable is learned, thus it maintains information from the entire distribution. In contrast, the MAP algorithm (i.e., equations \eqref{mapupdategamma}-\eqref{mapupdatetau}) updates each variable as the mode of the posterior distribution. With the uncertainties of all the other variables considered, each update step in the GHVI algorithm takes in more information than the MAP algorithm\cite{MLMurphy}. Therefore, the GHVI algorithm provides more accurate detection results than the MAP algorithm, as shown in the simulation results in Section \ref{sec:simulation}.
\subsubsection{Algorithm Time Complexity}
Bayesian learning methods are widely used in compressive sensing problems. However, they have high computation time complexity order due to the matrix inverse computation. Although \cite{zhang2021joint} introduced an inverse-free SBL algorithm via relaxed evidence lower bound in the circumstance of single-antenna APs, it is difficult to extend to multiple-antenna APs. The proposed GHVI not only models multi-antenna APs, but also avoids the matrix inverse by judiciously setting the mean-field in \eqref{meanviexpansion}. This results in only time complexity of $\mathcal{O}(IKNLM)$ where $I$ denotes the number of iterations. On the other hand, the complexity of the proposed MAP algorithm is $\mathcal{O}(IKN(LM+J))$, where $J$ denotes the number of iterations in gradient descent step in \eqref{mapupdateeta} and \eqref{mapetagradient}. Comparison of the complexity order of different algorithms is shown in Table \ref{tab_complexity}, where $T$ is the number of APs used to detect each user in the cell-free covariance-based method\cite{ganesan2021clustering}.
\begin{table*}[!t]
	\caption{Algorithm Complexity and Average Run Times (for the Setting of Fig. \ref{figcf_PMDPFA})\label{tab_complexity}}
	\centering
	\begin{tabular}{|c||c|c|}
		\hline
		Algorithm & Complexity & \thead{Average \\ Run Times (s)}\\
		\hline
		GHVI (proposed) & $\mathcal{O}(IKNLM)$ & 0.1555\\
		MAP (proposed) & $\mathcal{O}(IKN(LM+J))$ & 0.1607\\
		Cell-free covariance-based method\cite{ganesan2021clustering} & $\mathcal{O}\left(IN(T^3+KL^2+ML^2)\right)$ & 0.2591\\
		\makecell{Single-cell covariance-based method plus decision fusion\cite{xu2015active,covchen2022}} & $\mathcal{O}\left(IKNL^2\right)$ & 0.2029\\
		\makecell{Single-cell AMP plus LLR fusion\cite{dAMP,cranmp}} & $\mathcal{O}(IKNM^2)$ & 0.1495\\
		\makecell{Extension of single-cell GMMV-AMP\cite{gmmvamp} with LLR fusion} & $\mathcal{O}(IKNLM)$ & 0.1701\\
		\makecell{Single-cell Gaussian-gamma VI plus decision fusion\cite{crangg}} & $\mathcal{O}(IKML^3)$ & 0.1814\\
		\hline
	\end{tabular}
\end{table*}
\section{Simulation Results and Discussions}
\label{sec:simulation}
We evaluate the performance of the proposed MAP and GHVI methods, and compare them with five existing activity detection methods for cell-free systems:
\begin{itemize}
	\item Cell-free covariance-based method\cite{ganesan2021clustering}
	\item Single-cell covariance-based method with direct decision fusion\cite{xu2015active,covchen2022}
	\item Single-cell AMP with LLR fusion\cite{dAMP,cranmp}
	\item Extension of the single-cell GMMV-AMP\cite{gmmvamp} with an additional step of LLR fusion
	\item Single-cell Gaussian-gamma VI with direct decision fusion\cite{crangg}.
\end{itemize}
Notice that the cell-free covariance-based algorithms \cite{ganesan2021clustering} was designed to detect each user with only a small number of dominant APs ($\leq3$), we extended it such that it works with all available APs. The proposed algorithms and the competing methods with their associated properties are summarized in Table \ref{tab_algin}. All simulations are performed in Matlab version R2023a on a personal computer with Intel Core i7-12800H CPU and 16 GB RAM.
\subsection{Simulation Setting}
The simulation setup is as follows. In the cell-free system, there are $K=12$ APs with $M=8$ antennas in each AP and $N=200$ potential single-antenna devices that are assumed to be randomly located in a $3\text{ km}\times3\text{ km}$ area, and each device is active with probability $\varepsilon=0.1$ (unknown to the proposed algorithms). The pilot sequence length is set as $L=30$. The pathloss between user $n$ and the AP $k$ is $P_L=-128.1-36.7\log_{10}(d_{nk})+\Psi_{kn}$ in dB, where $d_{nk}$ is the distance between the user $n$ and AP $k$, and $\Psi_{kn}\sim\mathcal{N}(0,16)$ is the shadow fading. The small-scale channel component is Rayleigh fading or Rician fading. Considering the LoS component, the Rician fading can be expressed as
\begin{align}
	\label{ricianfading}
	\mathbf{g}_{kn}=&\sqrt{\dfrac{K_{\text{Rician}}^{kn}}{1+K_{\text{Rician}}^{kn}}}\bar{\mathbf{g}}_{kn}+\sqrt{\dfrac{1}{1+K_{\text{Rician}}^{kn}}}\mathcal{CN}\left(\mathbf{0},\mathbf{I}_M\right)\nonumber\\
	\sim&\mathcal{CN}\left(\sqrt{\dfrac{K_{\text{Rician}}^{kn}}{1+K_{\text{Rician}}^{kn}}}\bar{\mathbf{g}}_{kn},\dfrac{1}{1+K_{\text{Rician}}^{kn}}\mathbf{I}_M\right),
\end{align}
where $\bar{\mathbf{g}}_{kn}=\left[1~\exp\left(j\theta_{kn}\right)~\cdots~\exp\left(j\left(M-1\right)\theta_{kn}\right)\right]^T$ is the normalized LoS component with $\theta_{kn}$ uniformly chosen from $\left[0,2\pi\right]$, and $K_{\text{Rician}}^{kn}\sim\mathcal{U}(0,0.6)$ is the Rician factor. The maximum transmit power of each user is $23{\text{ dBm}}$, and the noise power is set as $-109\text{ dBm}$\cite{liu2018massive}. The large-scale fading coefficients are compensated by power allocation to give the same largest received $\text{SNR}=6\text{ dB}$ (among all the APs)\cite{ganesan2021clustering,dAMP}. The hyper-parameters are set as $c=d=\kappa_{2}=\eta_n^0=\psi_n^0=\lambda_n^0=10^{-6}$, $\kappa_{1}>-\lambda_n^0/2$. Each point in the figures is obtained by averaging over $10^5$ simulation trials.
\subsection{Comparison When Assumptions \textbf{A1}-\textbf{A4} Are Satisfied}
\begin{figure}[!t]
	\centering
	\includegraphics[width=3.8in]{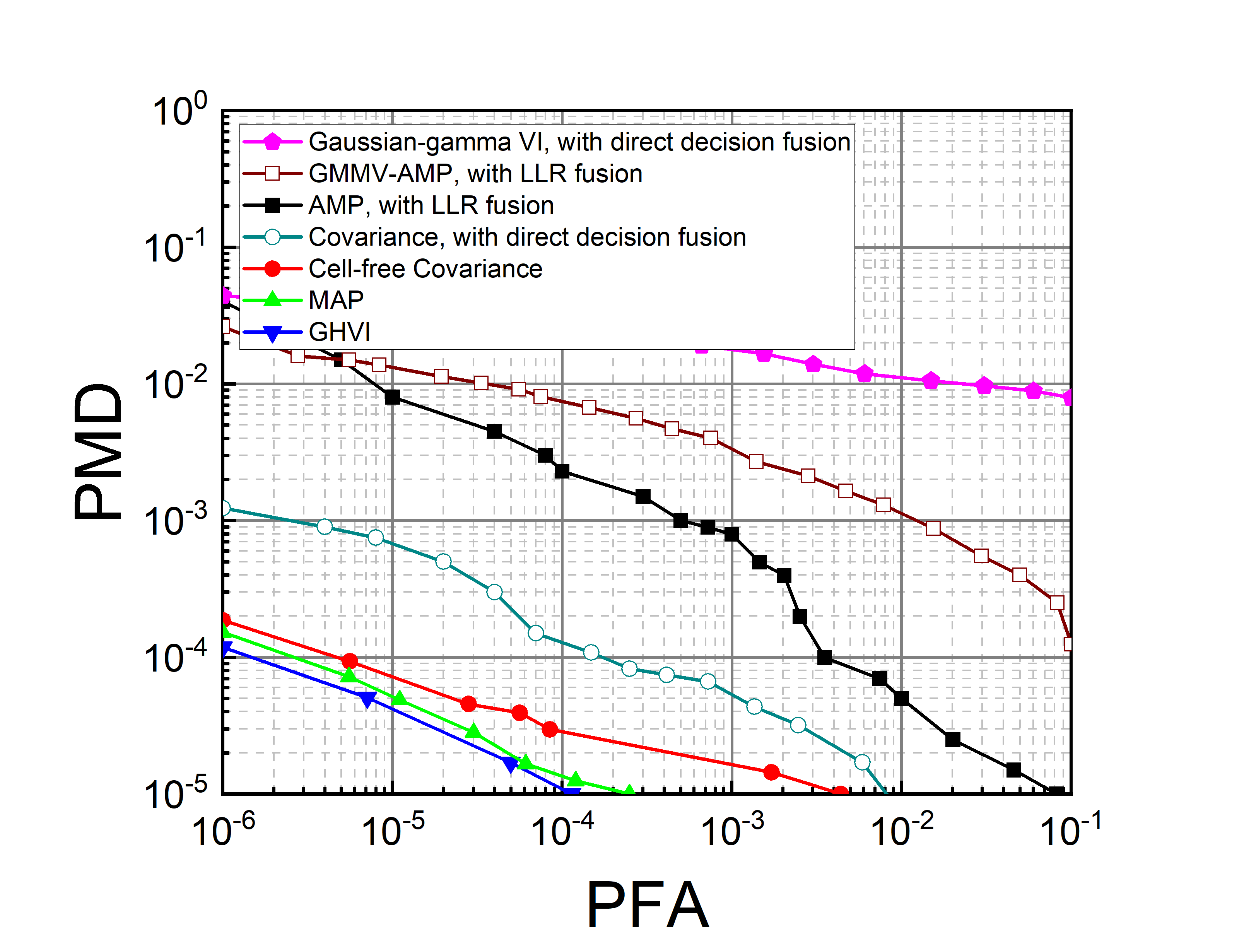}
	\caption{PMD versus PFA in cell-free system.}
	\label{figcf_PMDPFA}
\end{figure}

In Fig. \ref{figcf_PMDPFA}, we plot the probability of missed detection (PMD) versus the probability of false alarm (PFA) of different methods. For the two covariance-based methods and single-cell AMP with LLR fusion, the assumptions \textbf{A1}-\textbf{A4} are satisfied (i.e., all the large-scale fading coefficients $\left\{\gamma_{kn}\right\}_{k=1,n=1}^{K,N}$, noise variance $\sigma^2$, and user activity probability are precisely provided, and small-scale fading $\left\{\mathbf{g}_{kn}\right\}_{k=1,n=1}^{K,N}$ are Rayleigh fading $\mathcal{CN}(\mathbf{0},\mathbf{I}_M)$). This corresponds to a usual simulation setting in most existing works on activity detection. On the other hand, for the proposed MAP and GHVI algorithms, none of these information is known to the algorithms.

From Fig. \ref{figcf_PMDPFA}, we can observe that the single-cell Gaussian-gamma VI, AMP, GMMV-AMP, and covariance-based algorithms with fusion methods achieve a poor performance, because each AP makes its decisions independently. In contrast, the algorithms that impose activity status consistency (cell-free covariance-based method, the proposed MAP, and the proposed GHVI) achieve much lower error probabilities. Although the MAP algorithm does not fully utilize the uncertainty information from the probabilistic model, it already achieves impressive performance. This demonstrates the usefulness of the Bayesian model as a guide for introducing regularizations to learn all the unknown variables. Moreover, despite the absence of knowledge of system parameters, the proposed GHVI algorithm achieves unmistakably the lowest error probability among all the compared methods. Furthermore, the difference in performance between MAP and GHVI reveals that it is important to take into account the uncertainties of all the variables $\bm{\Theta}$ during inference, rather than simply obtaining a point estimate. Under this setting, the average run times of different methods are shown in the last column of Table \ref{tab_complexity}. We can observe that the proposed MAP and variational inference algorithms have comparable run times to those of AMP-based methods, while the covariance-based methods has the longest run time.

\begin{figure}[!t]
	\centering
	\includegraphics[width=3.8in]{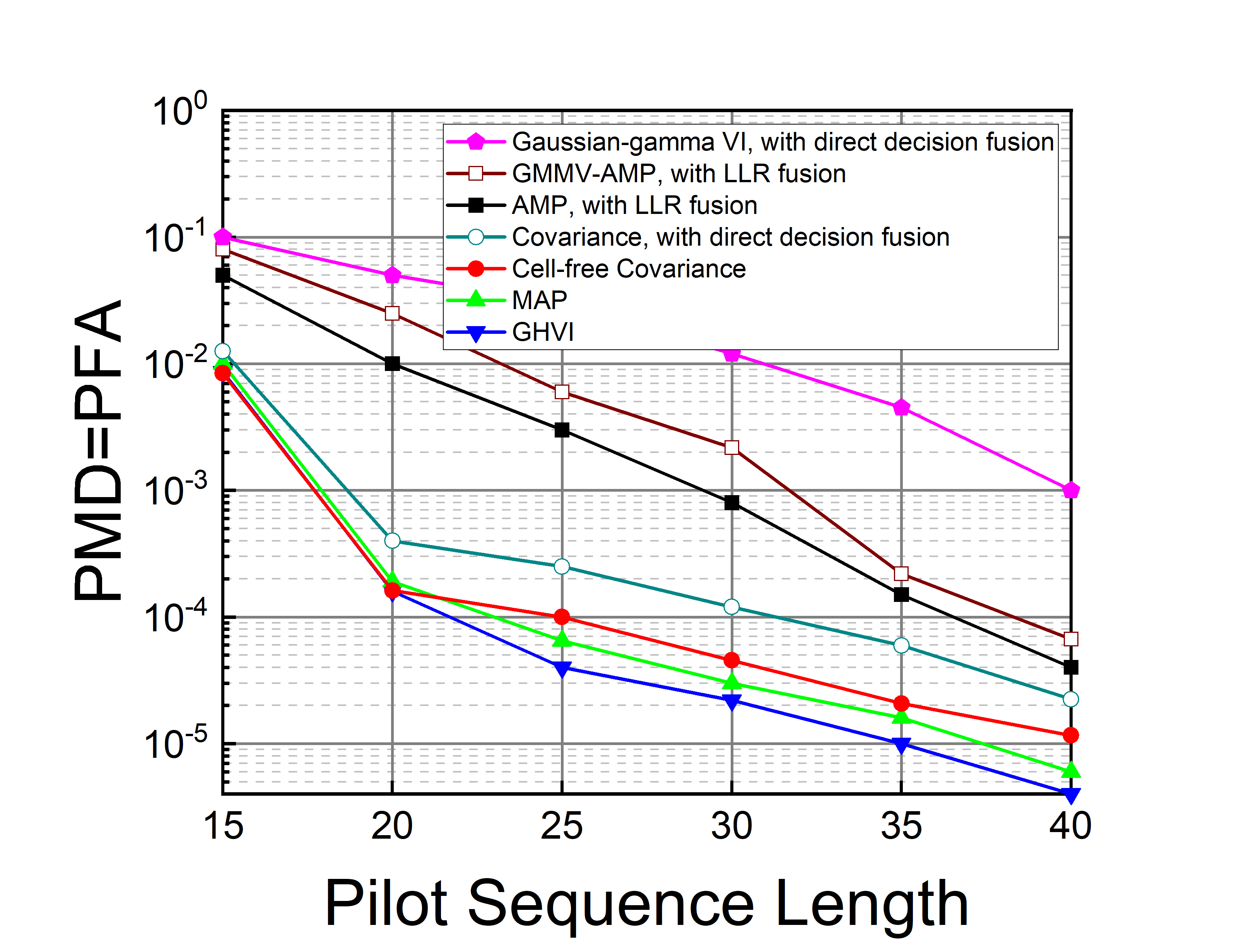}
	\caption{PMD=PFA versus the pilot sequence lengths.}
	\label{figcf_PMDPFAL}
\end{figure}

Fig. \ref{figcf_PMDPFAL} presents the performance of different methods under different pilot sequence lengths. Specifically, we compare the probability of error when the PMD is equal to the PFA by choosing a proper detection threshold. It is observed that the error probability of all the methods decreases with pilot sequence length because users are more easily distinguished through longer pilot sequence. By utilizing the consistent activity status, the cell-free covariance-based method, the proposed MAP, and the proposed GHVI can achieve reasonable performance even with very short pilot sequences. Moreover, with the regularizations introduced in the Bayesian model, the proposed MAP and GHVI perform the best under all considered pilot sequence lengths, even without any information from assumptions \textbf{A1}-\textbf{A4}.
\subsection{Comparison When Assumptions \textbf{A1}-\textbf{A4} Are Not Satisfied}
\begin{figure}[!t]
	\centering
	\includegraphics[width=3.8in]{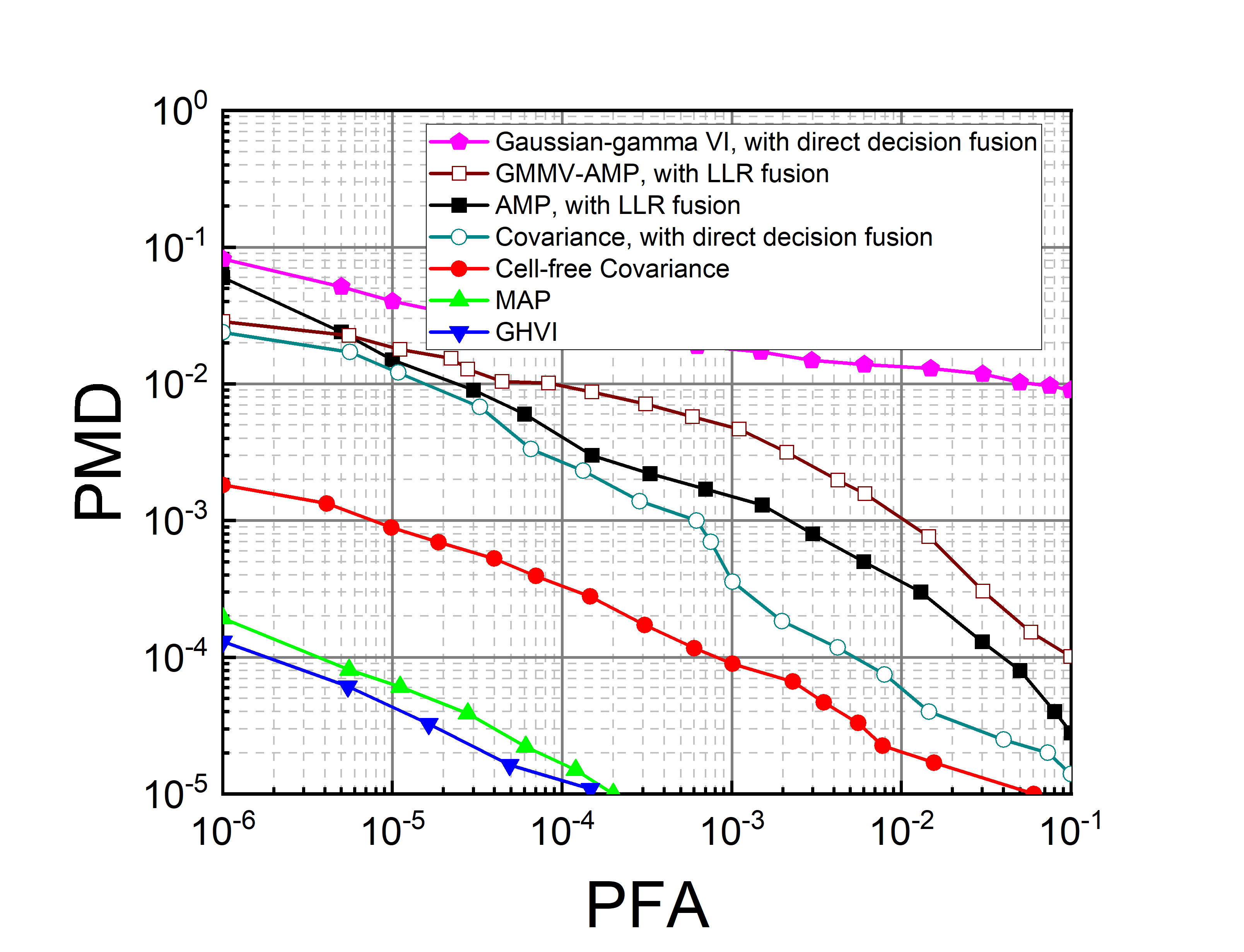}
	\caption{PMD versus PFA in cell-free system when assumptions \textbf{A1}-\textbf{A4} are violated.}
	\label{figcf_PMDPFAviolated}
\end{figure}

Figs. \ref{figcf_PMDPFA} and \ref{figcf_PMDPFAL} are obtained by assuming assumptions \textbf{A1}-\textbf{A4} are satisfied. In order to evaluate the performance of different algorithms under imprecise knowledge of the system parameters, in Fig. \ref{figcf_PMDPFAviolated}, we investigate the probability of error under violation of these assumptions. To be specific, in the simulations, each pathloss has an uncertainty drawn from uniform distribution $\mathcal{U}(0,2)$, 30\% of the devices have Rician fading, activity probability $\varepsilon\sim\mathcal{U}\left(0.1,0.2\right)$, and the noise power has an uncertainty $\mathcal{N}(0,0.2)$. For the AMP-based method, we input an activity probability $\epsilon=0.1$.

By comparing Figs. \ref{figcf_PMDPFA} and \ref{figcf_PMDPFAviolated}, we can see that cell-free covariance-based method, single-cell covariance-based method with direct detection fusion, and AMP with LLR fusion degrade significantly when the knowledge of system parameters is not accurate. For GMMV-VMP with LLR fusion and Gaussian-gamma VI with direct decision fusion, their performances in Fig. \ref{figcf_PMDPFA} and Fig. \ref{figcf_PMDPFAviolated} are similar (but at less than satisfactory accuracies), as they have the ability to estimate these system parameters. In contrast, by placing appropriate priors on various unknown parameters, the proposed Bayesian methods show almost the same impressively high detection accuracies in both Fig. \ref{figcf_PMDPFA} and Fig. \ref{figcf_PMDPFAviolated}. To investigate the degradation caused by violation of each assumption, in the following, we inspect the performance under violation of \textbf{A1}-\textbf{A4} one at a time.

\begin{figure}[!t]
	\centering
	\includegraphics[width=3.8in]{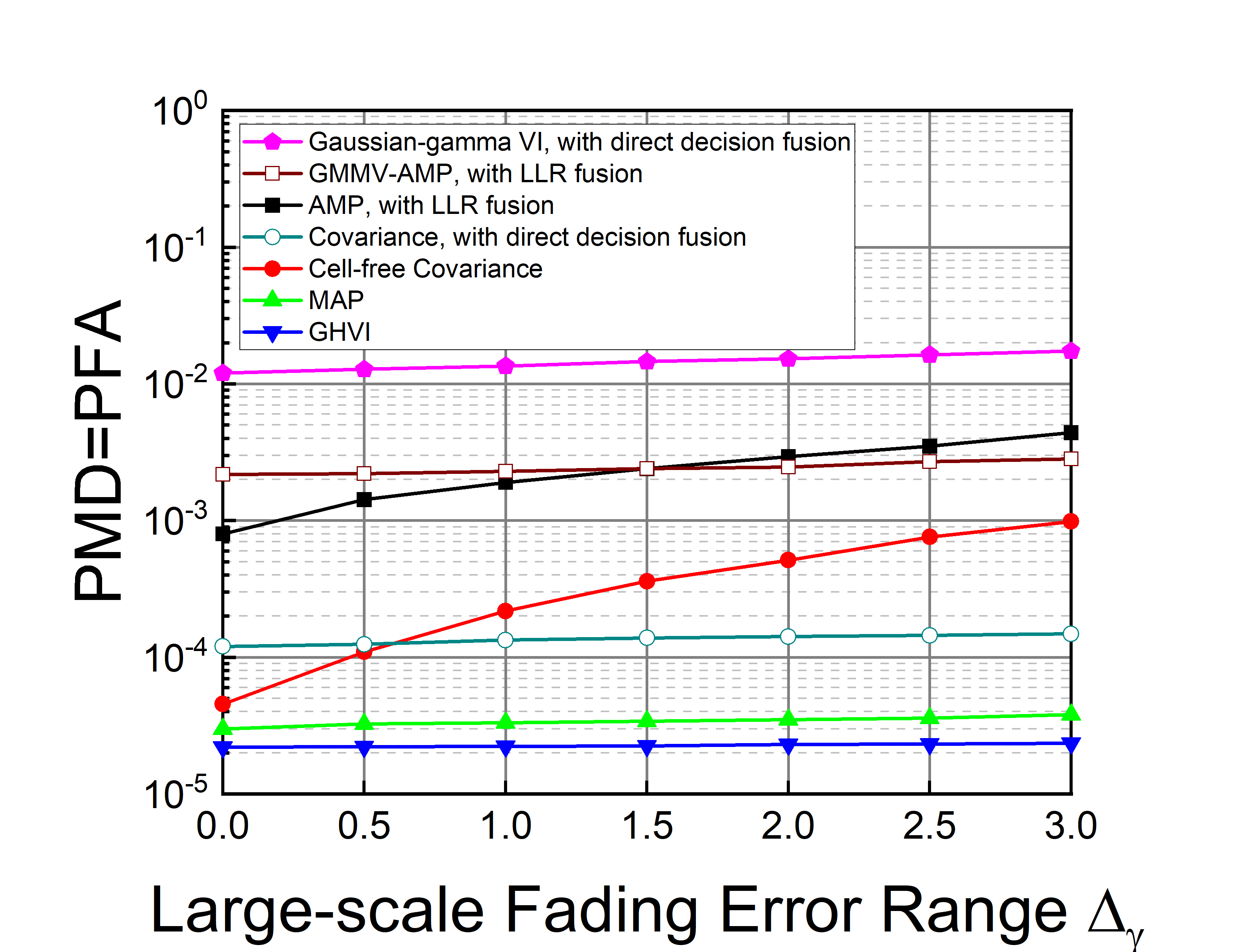}
	\caption{PMD=PFA versus the error in the large-scale fading coefficients.}
	\label{figcf_PMDPFAsigma}
\end{figure}

Firstly, for the violation of large-scale fading assumption (i.e., assumption \textbf{A1}), we add an unknown random component drawn from $\mathcal{U}(0,\Delta_{\gamma})$ to each of the pathloss $P_L$. Fig. \ref{figcf_PMDPFAsigma} shows the performance under different error levels with $\Delta_{\gamma}$ taking values from $0.5$ to $3$ in dB. With the true pathloss values around $-125\text{ dB}$, the introduced uncertainty corresponds to a relative error of large-scale fading coefficients ranges from 10\% to 50\%. We can observe that both the AMP method with LLR fusion and cell-free covariance-based method are profoundly affected even by minor large-scale fading uncertainty (e.g., $\Delta_{\gamma}=0.5$). This phenomenon is not surprising as the cell-free covariance-based method requires precise large-scale fading coefficients to determine the covariance matrix which directly affects the objective function (\ref{covarianceobj}). On the other hand, the minimum mean square error estimator in the AMP also depends on precise large-scale fading coefficients. In contrast, the proposed MAP and GHVI provide more robust performance, because the proposed Bayesian methods treat the large-scale fading coefficients as unknowns and learn their values from the observations. The Gaussian-gamma VI, GMMV-AMP, and covariance-based algorithms with fusion methods also learn the large-scale fading coefficients, therefore their detection performances stay almost constant with respect to large-scale fading error. However, they are not as good as the proposed methods because the consistent activity status is neglected. Furthermore, GHVI performs uniformly better than MAP estimator because it learns the variational distributions in (\ref{updatemugamma}) and (\ref{updatesigmagamma}) rather than a point estimate.

\begin{figure}[!t]
	\centering
	\includegraphics[width=3.8in]{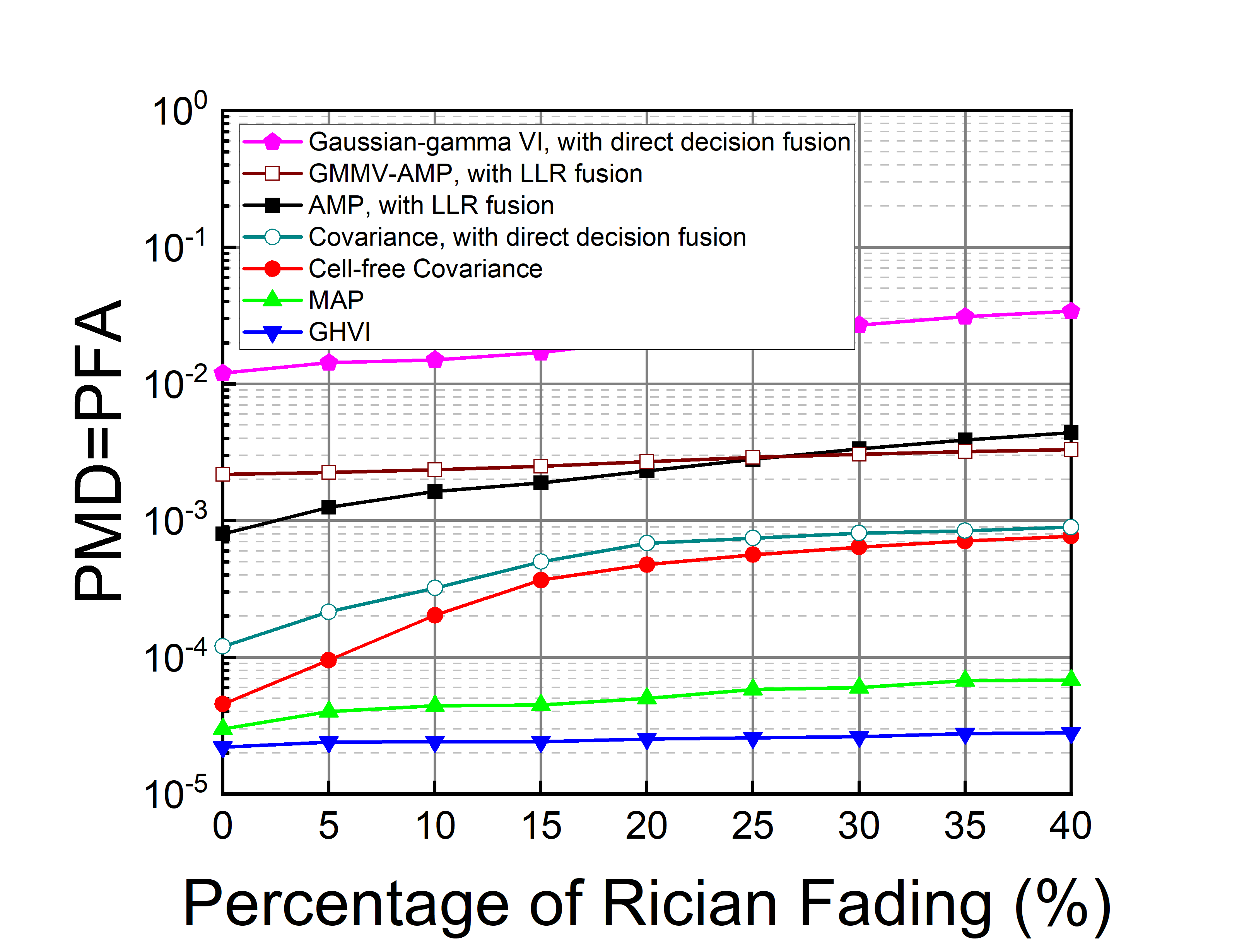}
	\caption{PMD=PFA versus the percentage of the users having Rician fading.}
	\label{figcf_PMDPFArician}
\end{figure}

Due to the potential LoS propagation, assumption \textbf{A2} may not be valid in practice. In order to investigate the effect of assumption \textbf{A2} violation, we randomly select a certain percentage of users (out of the 200 users) to have Rician fading channel. In Fig. \ref{figcf_PMDPFArician}, we show the performance under different percentages of devices having Rician fadings. We can see that the two covariance-based methods degrade severely when there exists LoS propagation, and the probability of error increases with the proportion of devices experiencing Rician fading. This is due to the fact that the covariance matrix in their objective function \eqref{covarianceobj} explicitly depends on the precise channel statistics, and more users experiencing Rician fading causes a significant mismatch between the assumed covariance matrix and that of the received signal. As a comparison, Gaussian-gamma VI and two AMP-based algorithms with fusion methods are less affected than the covariance-based method, because they estimate the small-scale fading even if there is LoS component. However, similar to Fig. \ref{figcf_PMDPFAsigma}, they still have higher error probabilities than the proposed MAP and GHVI due to the negligence of consistent activity status. Compared to the MAP, the variational distribution learned in GHVI contains more information about the channel statistics. This more sophisticated inference in turn leads to a better estimate of the parameters $z_n$, and results in the best performance among all algorithms.

\begin{figure}[!t]
	\centering
	\includegraphics[width=3.8in]{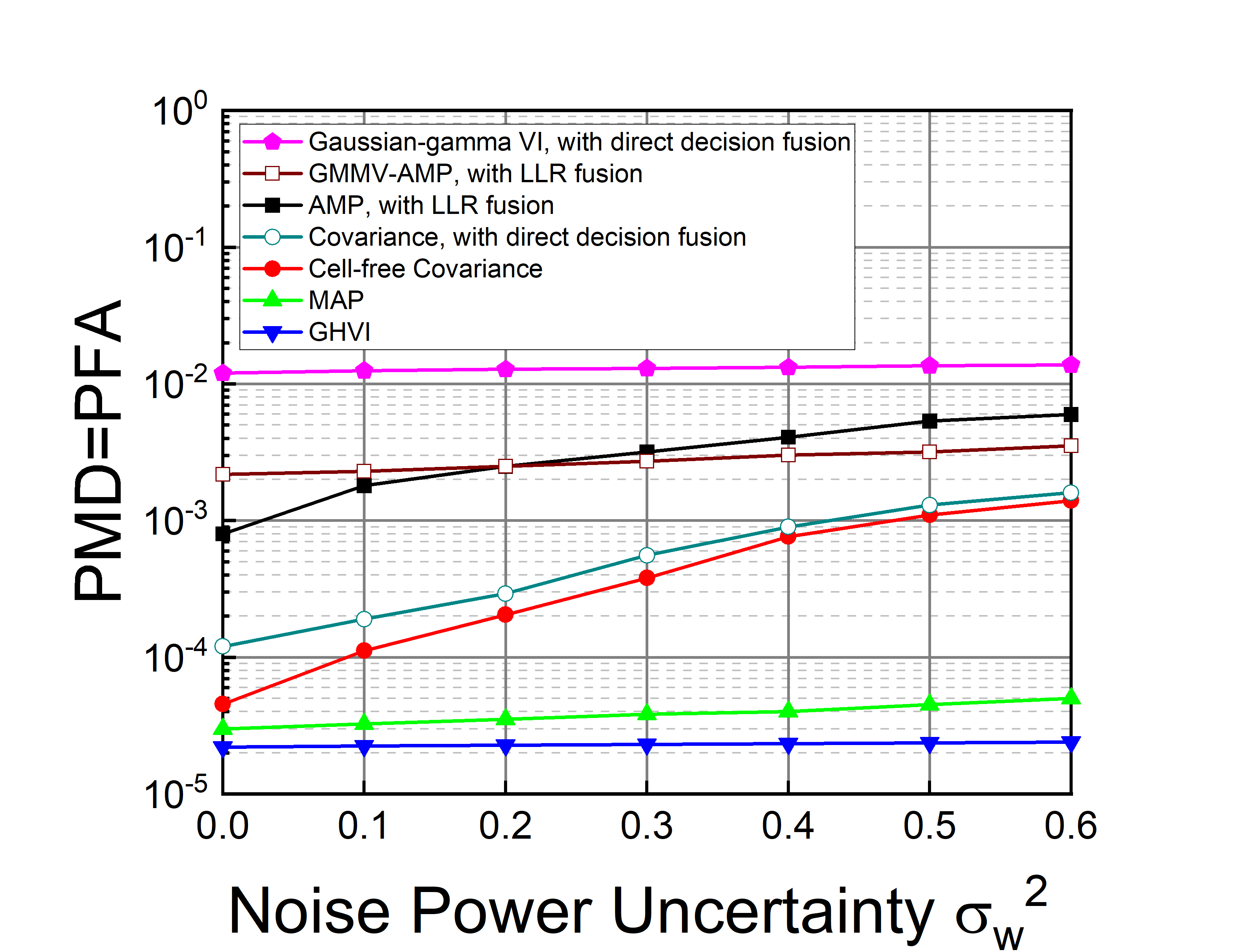}
	\caption{PMD=PFA versus the error uncertainty of noise power.}
	\label{figcf_PMDPFAtau}
\end{figure}

In Fig. \ref{figcf_PMDPFAtau}, we evaluate the performance of various methods when the actual noise power at each AP deviates from the assumed value by a random perturbation $\mathcal{N}\left(0,\sigma_{w}^2\right)$ in dBm, where $\sigma_{w}^2$ ranges from 0.1 to 0.6 (equivalently 1-10\% relative error). We can see that even a small error (0.1-0.2) can significantly degrade the performance of two covariance-based methods because their objective function (\ref{covarianceobj}) heavily relies on the precise knowledge of noise variance. The AMP algorithm is also heavily affected by imprecise noise power because its statistical model requires noise variance as a known parameter. Although the Gaussian-gamma VI and GMMV-AMP algorithms are less affected due to their abilities of estimating the noise power, they have relatively high error probabilities because the detection at each AP is independent. In comparison, the proposed Bayesian methods show robust and much better performance, because in addition to the noise precision estimation, the consistent activity status is utilized in the proposed Bayesian model. Similar to Figs. \ref{figcf_PMDPFAsigma} and \ref{figcf_PMDPFArician}, GHVI performs better than MAP algorithm because the learned uncertainties of various parameters in GHVI leads to more robust performance.

\begin{figure}[!t]
	\centering
	\includegraphics[width=3.8in]{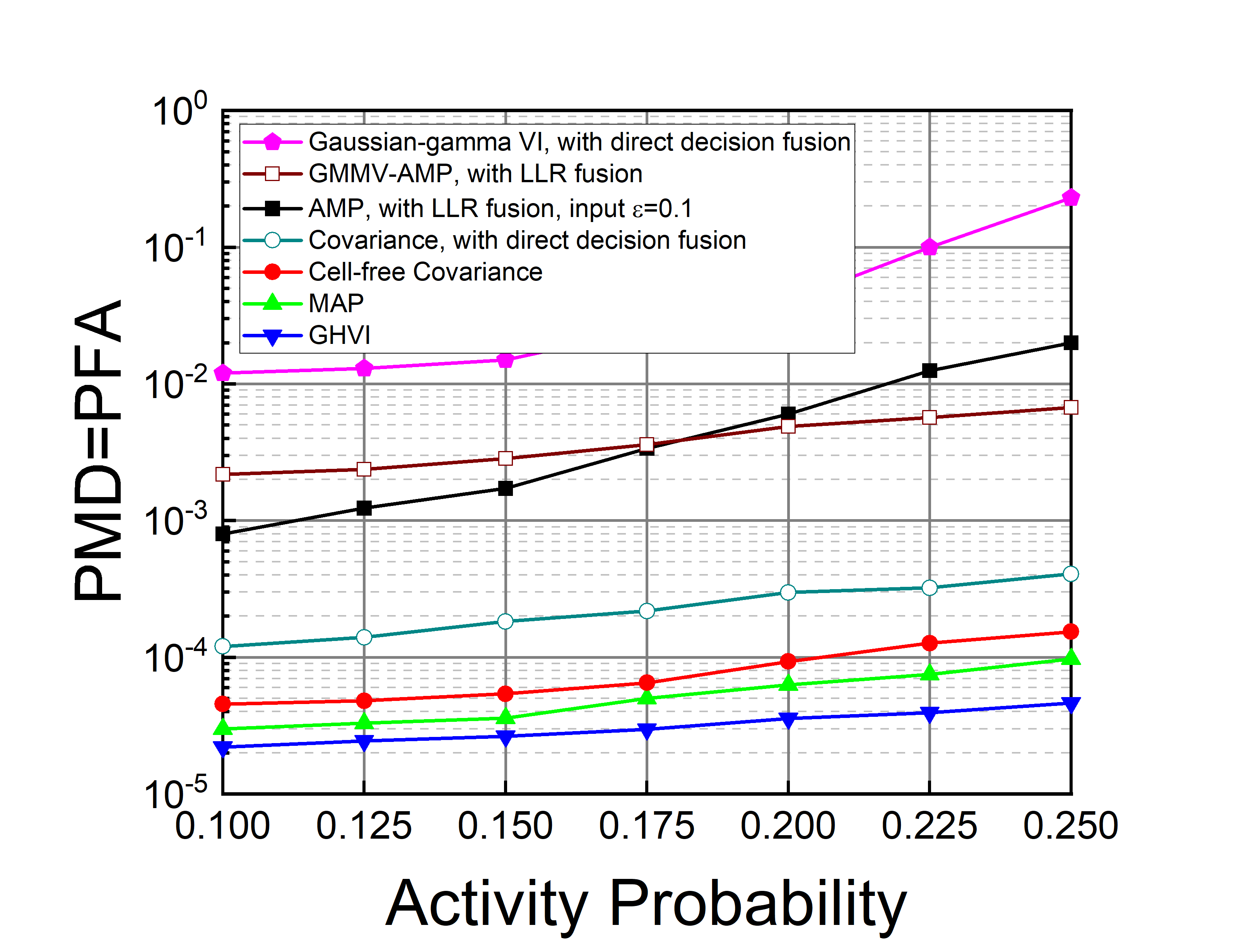}
	\caption{PMD=PFA versus the activity probability.}
	\label{figcf_PMDPFAAP}
\end{figure}

Finally, in Fig. \ref{figcf_PMDPFAAP}, we evaluate the performance under different activity probabilities. For the AMP method that requires the information about activity probability, we input a fixed $\varepsilon=0.1$. From Fig. \ref{figcf_PMDPFAAP}, we can observe that the AMP method degrades significantly when the activity probability is imprecise. In contrast, the proposed MAP and GHVI are more robust under a wide range of activity levels. The covariance-based methods and GMMV-AMP algorithms are also not affected by the activity probability too much, as they do not require activity probability as an input to the algorithms. It can be seen that the Gaussian-gamma prior performs well when the user activity level is around 10\%-15\%. However, when the user activity level further increases, it degrades considerably. This is because Gaussian-gamma prior is a traditional sparsity enhancing prior that was not designed to be flexible to learn different levels of sparsity. This shows that simply using Bayesian model and performing variational inference may not give satisfactory performance. The choice of a suitable prior distribution is also paramount in the robustness of the inference algorithm.
\section{Conclusions}
\label{sec:conclusion}
In this paper, a Bayesian activity detection model was proposed for cell-free massive connectivity wireless systems. By leveraging the flexible and sparsity-enhancing GH prior, the proposed probabilistic modeling is capable of adapting to different user activity levels under unknown large-scale fading coefficients, small-scale fading statistics, and noise variance at APs. Based on the proposed model, an MAP algorithm and a variational inference algorithm were derived to estimate the user activity together with these unknown system parameters. Extensive simulations showed that the proposed algorithms achieve better activity detection accuracy than state-of-the-art covariance-based and AMP-based methods, even in the best scenarios where all the required system parameters of covariance-based and AMP-based methods are precisely provided. Furthermore, since the proposed algorithms do not require any of these system parameters as input, they maintain robust detection performance even if there exists uncertainty in the system parameters while the competing algorithms degrade severely.
\appendix
\section{Derivations for Optimal Variational Distributions}
\subsection{The Logarithm of the Joint Distribution}
From \eqref{jointdistribution}, the logarithm of the joint distribution is
\begin{align}
	&\ln p(\mathcal{Y},\bm{\Theta})\nonumber\\
	=&KLM\ln\tau-\tau\sum_{k=1}^{K}\left\|\mathbf{Y}_k-\sum_{n=1}^{N}\gamma_{kn}\mathbf{s}_n\mathbf{g}_{kn}^T\right\|_F^2\nonumber
\end{align}
\begin{align}
	\label{logjoint}
	&+\sum_{k=1}^{K}\sum_{n=1}^{N}\left(\frac{\ln z_n^{-1}-z_n^{-1}\gamma_{kn}^2}{2}\right)+\sum_{k=1}^{K}\sum_{n=1}^{N}\left(-\left\|\mathbf{g}_{kn}\right\|_2^2\right)\nonumber\\
	&+\sum_{n=1}^{N}\left\{\dfrac{\lambda_n^0}{2}\ln\dfrac{\eta_n^0}{\psi_n^0}-\ln\left[2K_{\lambda_n^0}(\sqrt{\eta_n^0\psi_n^0})\right]+(\lambda_n^0-1)\ln z_n\right.\nonumber\\
	&-\left.\dfrac{1}{2}(\eta_n^0z_n+\psi_n^0z_n^{-1})\right\}+\sum_{n=1}^{N}\left[\left(\kappa_{1}-1\right)\ln \eta_n^0-\kappa_{2}\eta_n^0\right]\nonumber\\
	&+\left(c-1\right)\ln\tau-d\tau+\text{const}.
\end{align}

Then various optimal variational distributions $\left\{Q(\bm{\Theta})\right\}_{j=1}^J$ can be derived by \eqref{meanvisolution} with details presented below.
\subsection{Optimal Variational Distribution of $\left\{\gamma_{kn}\right\}_{k=1,n=1}^{K,N}$}
By only focusing on the terms related to $\gamma_{kn}$, we have
\begin{align}
	\label{derivationgamma}
	&\ln Q^*(\gamma_{kn})\nonumber\\
	\propto&\mathbb{E}_{Q(\bm{\Theta}\backslash\gamma_{kn})}\left[-\tau\left\|\mathbf{Y}_k-\sum_{n=1}^{N}\gamma_{kn}\mathbf{s}_n\mathbf{g}_{kn}^T\right\|_F^2-\dfrac{z_n^{-1}}{2}\gamma_{kn}^2\right]\nonumber\\
	&+\text{const}\nonumber\\
	=&\mathbb{E}\left\{-\tau\text{Tr}\left[(\mathbf{Y}_k-\sum_{m\neq n}\gamma_{km}\mathbf{s}_m\mathbf{g}_{km}^T-\gamma_{kn}\mathbf{s}_n\mathbf{g}_{kn}^T)^H\right.\right.\nonumber\\
	&\times\left.\left.(\mathbf{Y}_k-\sum_{m\neq n}\gamma_{km}\mathbf{s}_m\mathbf{g}_{km}^T-\gamma_n\mathbf{s}_n\mathbf{g}_{kn}^T)-\dfrac{z_n^{-1}}{2}\gamma_{kn}^2\right]\right\}\nonumber\\
	&+\text{const}\nonumber\\
	=&\mathbb{E}\left\{-\tau(\mathbf{s}_n^H\mathbf{s}_n)(\mathbf{g}_{kn}^H\mathbf{g}_{kn})\gamma_{kn}^2-\dfrac{z_n^{-1}}{2}\gamma_{kn}^2\right.\nonumber\\
	&+\left.\tau\text{Tr}\left[(\mathbf{Y}_k-\sum_{m\neq n}\gamma_{km}\mathbf{s}_m\mathbf{g}_{km}^T)^H(\gamma_{kn}\mathbf{s}_n\mathbf{g}_{kn}^T)\right.\right.\nonumber\\
	&\left.\left.+(\gamma_{kn}\mathbf{s}_n\mathbf{g}_{kn}^T)^H(\mathbf{Y}_k-\sum_{m\neq n}\gamma_{km}\mathbf{s}_m\mathbf{g}_{km}^T)\right]\right\}+\text{const}\nonumber\\
	=&2\langle\tau\rangle\text{Tr}\left\{\Re\left[\langle\mathbf{g}_{kn}^T\rangle^H\mathbf{s}_n^H\left(\mathbf{Y}_k-\sum_{m\neq n}\langle\gamma_{km}\rangle\mathbf{s}_m\langle\mathbf{g}_{km}^T\rangle\right)\right]\right\}\nonumber\\
	&\times\gamma_{kn}-\left[\langle\tau\rangle(\mathbf{s}_n^H\mathbf{s}_n)\langle\mathbf{g}_{kn}^H\mathbf{g}_{kn}\rangle-\langle \dfrac{z_n^{-1}}{2}\rangle\right]\gamma_{kn}^2+\text{const}.
\end{align}

By comparing \eqref{derivationgamma} to the form of Gaussian distribution, it can be inferred that the optimal $Q(\gamma_{kn})$ follows the Gaussian distribution $\mathcal{N}(\gamma_{kn}|\hat{\mu}_{kn}^{\gamma},\hat{\sigma}_{kn}^{\gamma})$ with $\hat{\mu}_{kn}^{\gamma}$ and $\hat{\sigma}_{kn}^{\gamma}$ given by \eqref{updatemugamma} and \eqref{updatesigmagamma}.
\subsection{Optimal Variational Distribution of $\left\{\mathbf{g}_{kn}\right\}_{k=1,n=1}^{K,N}$}
By only focusing on the terms related to $\mathbf{g}_{kn}$, we have
\begin{align}
	&\ln Q^*(\mathbf{g}_{kn})\nonumber\\
	\propto&\mathbb{E}_{Q(\bm{\Theta}\backslash\mathbf{g}_{kn})}\left[-\tau\left\|\mathbf{Y}_k-\sum_{n=1}^{N}\gamma_{kn}\mathbf{s}_n\mathbf{g}_{kn}^T\right\|_F^2-\left\|\mathbf{g}_{kn}\right\|_2^2\right]\nonumber
\end{align}
\begin{align}
	\label{derivationg}
	&+\text{const}\nonumber\\
	=&\mathbb{E}\left\{-\tau\text{Tr}\left[(\mathbf{Y}_k^T-\sum_{m\neq n}\gamma_{km}\mathbf{g}_{km}\mathbf{s}_m^T-\gamma_{kn}\mathbf{g}_{kn}\mathbf{s}_n^T)\right.\right.\nonumber\\
	&\times\left.\left.(\mathbf{Y}_k^T-\sum_{m\neq n}\gamma_{km}\mathbf{g}_{km}\mathbf{s}_m^T-\gamma_{kn}\mathbf{g}_{kn}\mathbf{s}_n^T)^H\right]-\mathbf{g}_{kn}^H\mathbf{g}_{kn}\right\}\nonumber\\
	&+\text{const}\nonumber\\
	=&\mathbb{E}\{\tau\text{Tr}\left[(\mathbf{Y}_k^T-\sum_{m\neq n}\gamma_{km}\mathbf{g}_{km}\mathbf{s}_m^T)\gamma_{kn}\bar{\mathbf{s}}_n\mathbf{g}_{kn}^H+\gamma_{kn}\mathbf{g}_{kn}\mathbf{s}_n^T\right.\nonumber\\
	&\times\left.(\mathbf{Y}_k^T-\sum_{m\neq n}\gamma_{km}\mathbf{g}_{km}\mathbf{s}_m^T)^H\right]-\tau\gamma_{kn}^2\mathbf{g}_{kn}^H(\mathbf{s}_n^H\mathbf{s}_n)\mathbf{g}_{kn}\nonumber\\
	&-\mathbf{g}_{kn}^H\mathbf{g}_{kn}\}+\text{const}\nonumber\\
	=&\mathbf{g}_{kn}^H\langle\tau\rangle\langle\gamma_{kn}\rangle\left(\mathbf{Y}_k^T-\sum_{m\neq n}\langle\gamma_{km}\rangle\langle\mathbf{g}_{km}\rangle\mathbf{s}_m^T\right)\bar{\mathbf{s}}_n\nonumber\\
	&+\langle\tau\rangle\langle\gamma_{kn}\rangle\mathbf{s}_n^T\left(\mathbf{Y}_k^T-\sum_{m\neq n}\langle\gamma_{km}\rangle\langle\mathbf{g}_{km}\rangle\mathbf{s}_m^T\right)^H\mathbf{g}_{kn}\nonumber\\
	&-\mathbf{g}_{kn}^H\left(\langle\tau\rangle\langle\gamma_{kn}^2\rangle\mathbf{s}_n^H\mathbf{s}_n+1\right)\mathbf{g}_{kn}+\text{const}.
\end{align}

By comparing \eqref{derivationg} to the form of complex Gaussian distribution, it can be inferred that the optimal $Q(\mathbf{g}_{kn})$ follows the complex Gaussian distribution $\mathcal{CN}(\mathbf{g}_{kn}|\hat{\mu}_{kn}^{\mathbf{g}},\hat{\Sigma}_{kn}^{\mathbf{g}})$ with $\hat{\mu}_{kn}^{\mathbf{g}}$ and $\hat{\Sigma}_{kn}^{\mathbf{g}}$ given by \eqref{updatemug} and \eqref{updatesigmag}.
\subsection{Optimal Variational Distribution of $\left\{z_n\right\}_{n=1}^N$}
By only focusing on the terms related to $z_n$, we have
\begin{align}
	\label{derivationz}
	&\ln Q^*(z_n)\nonumber\\
	\propto&\mathbb{E}_{Q(\bm{\Theta}\backslash z_n)}[\frac{1}{2}(K\ln z_n^{-1}-\sum_{k=1}^{K}z_n^{-1}\gamma_{kn}^2)+(\lambda_n^0-1)\ln z_n\nonumber\\
	&-\frac{1}{2}(\eta_n^0z_n+\psi_n^0z_n^{-1})]+\text{const}\nonumber\\
	=&-\frac{1}{2}\eta_n^0 z_n-\frac{1}{2}(\psi_n^0+\sum_{k=1}^{K}\langle\gamma_{kn}^2\rangle)z_n^{-1}+(\lambda_n^0-1-\frac{K}{2})\ln z_n\nonumber\\
	&+\text{const}.
\end{align}

By comparing \eqref{derivationz} to the form of GIG distribution, it can be inferred that the optimal $Q(z_n)$ follows the $\text{GIG}(z_n|\hat{\eta}_n,,\hat{\psi}_n,\hat{\lambda}_n)$ with $\hat{\eta}_n$, $\hat{\psi}_n$ and $\hat{\lambda}_n$ given by \eqref{updatea}-\eqref{updatelambda}.
\subsection{Optimal Variational Distribution of $\left\{\eta_n^0\right\}_{n=1}^N$}
By only focusing on the terms related to $\eta_n^0$, we have
\begin{align}
	\label{derivationa0}
	&\ln Q^*(\eta_n^0)\nonumber\\
	=&\mathbb{E}_{Q(\bm{\Theta}\backslash \eta_n^0)}\left[\left(\frac{\lambda_n^0}{2}\ln\frac{\eta_n^0}{\psi_n^0}-\frac{1}{2}\eta_n^0z_n\right)+(\kappa_{1}-1)\ln \eta_n^0\right.\nonumber\\
	&-\left.\kappa_{2}\eta_n^0\right]+\text{const}\nonumber\\
	=&\left(\kappa_{1}+\frac{\lambda_n^0}{2}-1\right)\ln \eta_n^0-\left(\kappa_{2}+\frac{\langle z_n\rangle}{2}\right)\eta_n^0+\text{const}.
\end{align}

By comparing \eqref{derivationa0} to the form of Gamma distribution, it can be inferred that the optimal $Q(\eta_n^0)$ follows the Gamma distribution $Ga(\eta_n^0|\hat{\kappa}_{n1},\hat{\kappa}_{n2})$ with $\hat{\kappa}_{n1}$ and $\hat{\kappa}_{n2}$ given by \eqref{updatek1} and \eqref{updatek2}.
\subsection{Optimal Variational Distribution of $\tau$}
By only focusing on the terms related to $\tau$, we have
\begin{align}
	\label{derivationtau}
	&\ln Q^*(\tau)\nonumber\\
	\propto&\mathbb{E}_{Q(\bm{\Theta}\backslash \tau)}[(KLM\ln\tau-\tau\sum_{k=1}^{K}|\mathbf{Y}_k-\sum_{n=1}^{N}\gamma_{kn}\mathbf{s}_n\mathbf{g}_{kn}^T\|_F^2\nonumber\\
	&+(c-1)\ln\tau-d\tau)]\nonumber\\
	=&(c+KLM-1)\ln\tau\nonumber\\
	&-\left(d+\sum_{k=1}^{K}\left\langle\left\|\mathbf{Y}_k-\sum_{n=1}^{N}\gamma_{kn}\mathbf{s}_n\mathbf{g}_{kn}^T\right\|_F^2\right\rangle\right)\tau.
\end{align}
It is easy to show $\tau$ also follows a Gamma distribution $Ga(\tau|\hat{c},\hat{d})$, with $\hat{c}$ given by \eqref{updatec} and $\hat{d}=d+\sum_{k=1}^{K}\left\langle\left\|\mathbf{Y}_k-\sum_{n=1}^{N}\gamma_{kn}\mathbf{s}_n\mathbf{g}_{kn}^T\right\|_F^2\right\rangle$. To compute $\hat{d}$, we expand the Frobenius norm:
\begin{align}
	\label{expandfnorm}
	&\sum_{k=1}^{K}\left\langle\left\|\mathbf{Y}_k-\sum_{n=1}^{N}\gamma_{kn}\mathbf{s}_n\mathbf{g}_{kn}^T\right\|_F^2\right\rangle\nonumber\\
	=&\sum_{k=1}^{K}\left\{\left\langle\sum_{n=1}^{N}\left[\gamma_{kn}^2\left(\mathbf{s}_n^H\mathbf{s}_n\right)\left(\mathbf{g}_{kn}^H\mathbf{g}_{kn}\right)\right]\right\rangle+\text{Tr}(\mathbf{Y}_k^H\mathbf{Y}_k)\right.\nonumber\\
	&+\text{Tr}\left[\sum_{n=1}^{N}\sum_{m\neq n}^{N}\langle\gamma_{kn}\rangle\langle\gamma_{km}\rangle\left(\mathbf{s}_n\langle\mathbf{g}_{kn}^T\rangle\right)^H\left(\mathbf{s}_m\langle\mathbf{g}_{km}^T\rangle\right)\right]\nonumber\\
	&-\text{Tr}\left[\mathbf{Y}_k^H\left(\sum_{n=1}^{N}\langle\gamma_{kn}\rangle\mathbf{s}_n\langle\mathbf{g}_{kn}^T\rangle\right)\right.\nonumber\\
	&\left.\left.+\left(\sum_{n=1}^{N}\langle\gamma_{kn}\rangle\mathbf{s}_n\langle\mathbf{g}_{kn}^T\rangle\right)^H\mathbf{Y}_k\right]\right\},
\end{align}
where the first term can be expanded as
\begin{align}
	\label{expandsquaredmean}
	&\sum_{k=1}^K\sum_{n=1}^{N}\left\langle\left[\gamma_{kn}^2\left(\mathbf{s}_n^H\mathbf{s}_n\right)\left(\mathbf{g}_{kn}^H\mathbf{g}_{kn}\right)\right]\right\rangle\nonumber\\
	=&\sum_{k=1}^{K}\sum_{n=1}^{N}\left\{\left[\left(\mathbf{s}_n^H\mathbf{s}_n\right)\left(\langle\gamma_{kn}\rangle^2\langle\mathbf{g}_{kn}^H\rangle\langle\mathbf{g}_{kn}\rangle\right)\right]+\left(\mathbf{s}_n^H\mathbf{s}_n\right)\right.\nonumber\\
	&\times\left.\left[\hat{\sigma}_{kn}^{\gamma}\langle\mathbf{g}_{kn}^H\rangle\langle\mathbf{g}_{kn}\rangle+\langle\gamma_{kn}\rangle^2\text{Tr}\left(\hat{\Sigma}_{kn}^{\mathbf{g}}\right)+\hat{\sigma}_{kn}^{\gamma}\hat{\Sigma}_{kn}^{\mathbf{g}}\right]\right\}.
\end{align}

By putting \eqref{expandsquaredmean} into \eqref{expandfnorm}, it is eventually computed as
\begin{align}
	&\sum_{k=1}^K\left\langle\left\|\mathbf{Y}_k-\sum_{n=1}^{N}\gamma_{kn}\mathbf{s}_n\mathbf{g}_{kn}^T\right\|_F^2\right\rangle\nonumber\\
	=&\sum_{k=1}^{K}\left\{\text{Tr}(\mathbf{Y}_k^H\mathbf{Y}_k)+\sum_{n=1}^{N}\sum_{m=1}^{N}\text{Tr}\left[\langle\gamma_{kn}\rangle\langle\gamma_{km}\rangle\left(\mathbf{s}_n\langle\mathbf{g}_{kn}^T\rangle\right)^H\right.\right.\nonumber
\end{align}
\begin{align}
	\label{calculatetau}
	&\left.\times\left(\mathbf{s}_m\langle\mathbf{g}_{km}^T\rangle\right)\right]-\text{Tr}\left[\mathbf{Y}_k^H\left(\sum_{n=1}^{N}\langle\gamma_{kn}\rangle\mathbf{s}_n\langle\mathbf{g}_{kn}^T\rangle\right)\right.\nonumber\\
	&\left.\left.+\left(\sum_{n=1}^{N}\langle\gamma_{kn}\rangle\mathbf{s}_n\langle\mathbf{g}_{kn}^T\rangle\right)^H\mathbf{Y}_k\right]\right\}+\sum_{k=1}^K\sum_{n=1}^{N}\left(\mathbf{s}_n^H\mathbf{s}_n\right)\nonumber\\
	&\times\left[\hat{\sigma}_{kn}^{\gamma}\langle\mathbf{g}_{kn}^H\rangle\langle\mathbf{g}_{kn}\rangle+\langle\gamma_{kn}\rangle^2\text{Tr}\left(\hat{\Sigma}_{kn}^{\mathbf{g}}\right)+\hat{\sigma}_{kn}^{\gamma}\hat{\Sigma}_{kn}^{\mathbf{g}}\right]\nonumber\\
	=&\sum_{k=1}^{K}\sum_{n=1}^{N}\left\{\left(\mathbf{s}_n^H\mathbf{s}_n\right)\left[\hat{\sigma}_{kn}^{\gamma}\langle\mathbf{g}_{kn}^H\rangle\langle\mathbf{g}_{kn}\rangle+\langle\gamma_{kn}\rangle^2\text{Tr}\left(\hat{\Sigma}_{kn}^{\mathbf{g}}\right)\right.\right.\nonumber\\
	&\left.\left.+\hat{\sigma}_{kn}^{\gamma}\hat{\Sigma}_{kn}^{\mathbf{g}}\right]\right\}+\sum_{k=1}^K\left\|\mathbf{Y}_k-\sum_{n=1}^{N}\langle\gamma_{kn}\rangle\mathbf{s}_n\langle\mathbf{g}_{kn}^T\rangle\right\|_F^2,
\end{align}
which is the expression in \eqref{updated}.
\bibliography{references}

\begin{thebibliography}{10}
\providecommand{\url}[1]{#1}
\csname url@samestyle\endcsname
\providecommand{\newblock}{\relax}
\providecommand{\bibinfo}[2]{#2}
\providecommand{\BIBentrySTDinterwordspacing}{\spaceskip=0pt\relax}
\providecommand{\BIBentryALTinterwordstretchfactor}{4}
\providecommand{\BIBentryALTinterwordspacing}{\spaceskip=\fontdimen2\font plus
\BIBentryALTinterwordstretchfactor\fontdimen3\font minus
  \fontdimen4\font\relax}
\providecommand{\BIBforeignlanguage}[2]{{%
\expandafter\ifx\csname l@#1\endcsname\relax
\typeout{** WARNING: IEEEtran.bst: No hyphenation pattern has been}%
\typeout{** loaded for the language `#1'. Using the pattern for}%
\typeout{** the default language instead.}%
\else
\language=\csname l@#1\endcsname
\fi
#2}}
\providecommand{\BIBdecl}{\relax}
\BIBdecl

\bibitem{icassp2024}
H.~Zhang, Q.~Lin, Y.~Li, L.~Cheng, and Y.-C. Wu, ``Bayesian activity detection
  for massive connectivity in cell-free {IoT} networks,'' submitted to 2024
  IEEE International Conference on Acoustics, Speech, and Signal Processing
  (ICASSP 2024).

\bibitem{zhou2005}
Y.~Zhou, J.~Wang, and M.~Sawahashi, ``Downlink transmission of broadband
  {OFCDM} systems-part {I}: hybrid detection,'' \emph{IEEE Transactions on
  Communications}, vol.~53, no.~4, pp. 718--729, 2005,
  doi:10.1109/TCOMM.2005.844962.

\bibitem{liu2018massive}
L.~Liu and W.~Yu, ``Massive connectivity with massive {MIMO}—part {I}: Device
  activity detection and channel estimation,'' \emph{IEEE Transactions on
  Signal Processing}, vol.~66, no.~11, pp. 2933--2946, 2018,
  doi:10.1109/TSP.2018.2818082.

\bibitem{liu2018massive3}
L.~Liu, E.~G. Larsson, W.~Yu, P.~Popovski, C.~Stefanovic, and E.~de~Carvalho,
  ``Sparse signal processing for grant-free massive connectivity: A future
  paradigm for random access protocols in the internet of things,'' \emph{IEEE
  Signal Processing Magazine}, vol.~35, no.~5, pp. 88--99, 2018,
  doi:10.1109/MSP.2018.2844952.

\bibitem{schepker2013exploiting}
H.~F. Schepker, C.~Bockelmann, and A.~Dekorsy, ``Exploiting sparsity in channel
  and data estimation for sporadic multi-user communication,'' in \emph{ISWCS
  2013; The Tenth International Symposium on Wireless Communication Systems},
  2013, pp. 1--5.

\bibitem{wunder2014compressive}
G.~Wunder, P.~Jung, and C.~Wang, ``Compressive random access for post-lte
  systems,'' in \emph{2014 IEEE International Conference on Communications
  Workshops (ICC)}, 2014, pp. 539--544, doi:10.1109/ICCW.2014.6881254.

\bibitem{wunder2015compressive}
G.~Wunder, P.~Jung, and M.~Ramadan, ``Compressive random access using a common
  overloaded control channel,'' in \emph{2015 IEEE Globecom Workshops (GC
  Wkshps)}, 2015, pp. 1--6, doi:10.1109/GLOCOMW.2015.7414186.

\bibitem{schepker2012compressive}
H.~F. Schepker and A.~Dekorsy, ``Compressive sensing multi-user detection with
  block-wise orthogonal least squares,'' in \emph{2012 IEEE 75th Vehicular
  Technology Conference (VTC Spring)}, 2012, pp. 1--5,
  doi:10.1109/VETECS.2012.6240301.

\bibitem{applebaum2012asynchronous}
\BIBentryALTinterwordspacing
L.~Applebaum, W.~U. Bajwa, M.~F. Duarte, and R.~Calderbank, ``Asynchronous
  code-division random access using convex optimization,'' \emph{Physical
  Communication}, vol.~5, no.~2, pp. 129--147, 2012,
  doi:https://doi.org/10.1016/j.phycom.2011.09.006. [Online]. Available:
  \url{https://www.sciencedirect.com/science/article/pii/S187449071100053X}
\BIBentrySTDinterwordspacing

\bibitem{li2019activity}
Y.~Li, M.~Xia, and Y.-C. Wu, ``Activity detection for massive connectivity
  under frequency offsets via first-order algorithms,'' \emph{IEEE Transactions
  on Wireless Communications}, vol.~18, no.~3, pp. 1988--2002, 2019,
  doi:10.1109/TW.

\bibitem{8437359}
S.~Haghighatshoar, P.~Jung, and G.~Caire, ``Improved scaling law for activity
  detection in massive {MIMO} systems,'' in \emph{2018 IEEE International
  Symposium on Information Theory (ISIT)}, 2018, pp. 381--385,
  doi:10.1109/ISIT.2018.8437359.

\bibitem{9691883}
Z.~Wang, Y.-F. Liu, and L.~Liu, ``Covariance-based joint device activity and
  delay detection in asynchronous m{MTC},'' \emph{IEEE Signal Processing
  Letters}, vol.~29, pp. 538--542, 2022, doi:10.1109/LSP.2022.3144853.

\bibitem{liyangtransformer}
Y.~Li, Z.~Chen, Y.~Wang, C.~Yang, B.~Ai, and Y.-C. Wu, ``Heterogeneous
  transformer: A scale adaptable neural network architecture for device
  activity detection,'' \emph{IEEE Transactions on Wireless Communications},
  vol.~22, no.~5, pp. 3432--3446, 2023.

\bibitem{wunder2015sparse}
G.~Wunder, H.~Boche, T.~Strohmer, and P.~Jung, ``Sparse signal processing
  concepts for efficient {5G} system design,'' \emph{IEEE Access}, vol.~3, pp.
  195--208, 2015, doi:10.1109/ACCESS.2015.2407194.

\bibitem{lin2023}
Q.~Lin, Y.~Li, and Y.-C. Wu, ``Sparsity constrained joint activity and data
  detection for massive access: A difference-of-norms penalty framework,''
  \emph{IEEE Transactions on Wireless Communications}, vol.~22, no.~3, pp.
  1480--1494, 2023, doi:10.1109/TWC.2022.3204786.

\bibitem{zhu2010exploiting}
H.~Zhu and G.~B. Giannakis, ``Exploiting sparse user activity in multiuser
  detection,'' \emph{IEEE Transactions on Communications}, vol.~59, no.~2, pp.
  454--465, 2011, doi:10.1109/TCOMM.2011.121410.090570.

\bibitem{hannak2015joint}
G.~Hannak, M.~Mayer, A.~Jung, G.~Matz, and N.~Goertz, ``Joint channel
  estimation and activity detection for multiuser communication systems,'' in
  \emph{2015 IEEE International Conference on Communication Workshop (ICCW)},
  2015, pp. 2086--2091, doi:10.1109/ICCW.2015.7247489.

\bibitem{chen2017massive}
Z.~Chen and W.~Yu, ``Massive device activity detection by approximate message
  passing,'' in \emph{2017 IEEE International Conference on Acoustics, Speech
  and Signal Processing (ICASSP)}, 2017, pp. 3514--3518,
  doi:10.1109/ICASSP.2017.7952810.

\bibitem{chen2018sparse}
Z.~Chen, F.~Sohrabi, and W.~Yu, ``Sparse activity detection for massive
  connectivity,'' \emph{IEEE Transactions on Signal Processing}, vol.~66,
  no.~7, pp. 1890--1904, 2018, doi:10.1109/TSP.2018.2795540.

\bibitem{ding2019sparsity}
T.~Ding, X.~Yuan, and S.~C. Liew, ``Sparsity learning-based multiuser detection
  in grant-free massive-device multiple access,'' \emph{IEEE Transactions on
  Wireless Communications}, vol.~18, no.~7, pp. 3569--3582, 2019,
  doi:10.1109/TWC.2019.2915955.

\bibitem{ahn2019ep}
J.~Ahn, B.~Shim, and K.~B. Lee, ``{EP}-based joint active user detection and
  channel estimation for massive machine-type communications,'' \emph{IEEE
  Transactions on Communications}, vol.~67, no.~7, pp. 5178--5189, 2019,
  doi:10.1109/TCOMM.2019.2907853.

\bibitem{zhang2021joint}
X.~Zhang, F.~Labeau, L.~Hao, and J.~Liu, ``Joint active user detection and
  channel estimation via {Bayesian} learning approaches in {MTC}
  communications,'' \emph{IEEE Transactions on Vehicular Technology}, vol.~70,
  no.~6, pp. 6222--6226, 2021, doi:10.1109/TVT.2021.3077569.

\bibitem{zhang2021low}
S.~Zhang, Y.~Wang, and W.~Zhou, ``A low-complexity variational {Bayesian}
  learning algorithm on channel estimation using group sparse structure,'' in
  \emph{2021 7th International Conference on Computer and Communications
  (ICCC)}, 2021, pp. 245--250, doi:10.1109/ICCC54389.2021.9674470.

\bibitem{wang2022double}
Y.~Wang, Z.~Qiu, S.~Zhang, H.~Tian, and W.~Zhou, ``Double sparsity-based joint
  active user detection and channel estimation for m{MTC}-enabled massive
  {MIMO},'' in \emph{ICC 2022 - IEEE International Conference on
  Communications}, 2022, pp. 968--973, doi:10.1109/ICC45855.2022.9838652.

\bibitem{interdonato2019ubiquitous}
G.~Interdonato, E.~Bj{\"o}rnson, H.~Quoc~Ngo, P.~Frenger, and E.~G. Larsson,
  ``Ubiquitous cell-free massive {MIMO} communications,'' \emph{EURASIP Journal
  on Wireless Communications and Networking}, vol. 2019, no.~1, pp. 1--13,
  2019, doi:https://doi.org/10.1186/s13638-019-1507-0.

\bibitem{ngo2017cell}
H.~Q. Ngo, A.~Ashikhmin, H.~Yang, E.~G. Larsson, and T.~L. Marzetta,
  ``Cell-free massive {MIMO} versus small cells,'' \emph{IEEE Transactions on
  Wireless Communications}, vol.~16, no.~3, pp. 1834--1850, 2017,
  doi:10.1109/TWC.2017.2655515.

\bibitem{zhou2018}
L.~Liu, Y.~Zhou, V.~Garcia, L.~Tian, and J.~Shi, ``Load aware joint {CoMP}
  clustering and inter-cell resource scheduling in heterogeneous ultra dense
  cellular networks,'' \emph{IEEE Transactions on Vehicular Technology},
  vol.~67, no.~3, pp. 2741--2755, 2018, doi:10.1109/TVT.2017.2773640.

\bibitem{zhou2014}
V.~Garcia, Y.~Zhou, and J.~Shi, ``Coordinated multipoint transmission in dense
  cellular networks with user-centric adaptive clustering,'' \emph{IEEE
  Transactions on Wireless Communications}, vol.~13, no.~8, pp. 4297--4308,
  2014, doi:10.1109/TWC.2014.2316500.

\bibitem{tan2021}
F.~Tan, P.~Wu, Y.-C. Wu, and M.~Xia, ``Energy-efficient non-orthogonal
  multicast and unicast transmission of cell-free massive {MIMO} systems with
  {SWIPT},'' \emph{IEEE Journal on Selected Areas in Communications}, vol.~39,
  no.~4, pp. 949--968, 2021, doi:10.1109/JSAC.2020.3020110.

\bibitem{xu2015active}
X.~Xu, X.~Rao, and V.~K. Lau, ``Active user detection and channel estimation in
  uplink {CRAN} systems,'' in \emph{2015 IEEE International Conference on
  Communications (ICC)}, 2015, pp. 2727--2732, doi:10.1109/ICC.2015.7248738.

\bibitem{liyangdcov}
Y.~Li, Q.~Lin, Y.-F. Liu, B.~Ai, and Y.-C. Wu, ``Asynchronous activity
  detection for cell-free massive {MIMO}: From centralized to distributed
  algorithms,'' \emph{IEEE Transactions on Wireless Communications}, vol.~22,
  no.~4, pp. 2477--2492, 2023, doi:10.1109/TWC.2022.3211967.

\bibitem{crangg}
J.~Wang, J.~Yi, R.~Han, L.~Bai, and J.~Choi, ``Variational {Bayesian} inference
  for channel estimation and user activity detection in {C-RAN},'' \emph{IEEE
  Wireless Communications Letters}, vol.~9, no.~7, pp. 953--956, 2020,
  doi:10.1109/LWC.2020.2975785.

\bibitem{guo2021joint}
M.~Guo and M.~C. Gursoy, ``Joint activity detection and channel estimation in
  cell-free massive {MIMO} networks with massive connectivity,'' \emph{IEEE
  Transactions on Communications}, vol.~70, no.~1, pp. 317--331, 2022,
  doi:10.1109/TCOMM.2021.3122471.

\bibitem{guo2020sparse}
M.~Guo, M.~C. Gursoy, and P.~K. Varshney, ``Sparse activity detection in
  cell-free massive {MIMO} systems,'' in \emph{2020 IEEE International
  Symposium on Information Theory (ISIT)}, 2020, pp. 1177--1182,
  doi:10.1109/ISIT44484.2020.9174169.

\bibitem{guo2019distributed}
M.~Guo and M.~C. Gursoy, ``Distributed sparse activity detection in cell-free
  massive {MIMO} systems,'' in \emph{2019 IEEE Global Conference on Signal and
  Information Processing (GlobalSIP)}, 2019, pp. 1--5,
  doi:10.1109/GlobalSIP45357.2019.8969500.

\bibitem{multicell}
Z.~Chen, F.~Sohrabi, and W.~Yu, ``Multi-cell sparse activity detection for
  massive random access: Massive {MIMO} versus cooperative {MIMO},'' \emph{IEEE
  Transactions on Wireless Communications}, vol.~18, no.~8, pp. 4060--4074,
  2019, doi:10.1109/TWC.2019.2920823.

\bibitem{dAMP}
J.~Bai and E.~G. Larsson, ``Activity detection in distributed {MIMO}:
  Distributed {AMP} via likelihood ratio fusion,'' \emph{IEEE Wireless
  Communications Letters}, vol.~11, no.~10, pp. 2200--2204, 2022,
  doi:10.1109/LWC.2022.3197053.

\bibitem{cranmp}
Y.~Chi, L.~Liu, G.~Song, C.~Yuen, Y.~L. Guan, and Y.~Li, ``Message passing in
  {C-RAN}: Joint user activity and signal detection,'' in \emph{GLOBECOM 2017 -
  2017 IEEE Global Communications Conference}, 2017, pp. 1--6,
  doi:10.1109/GLOCOM.2017.8254230.

\bibitem{ganesan2021clustering}
U.~K. Ganesan, E.~Björnson, and E.~G. Larsson, ``Clustering-based activity
  detection algorithms for grant-free random access in cell-free massive
  {MIMO},'' \emph{IEEE Transactions on Communications}, vol.~69, no.~11, pp.
  7520--7530, 2021, doi:10.1109/TCOMM.2021.3102635.

\bibitem{lsfc}
C.~Wang, O.~Y. Bursalioglu, H.~Papadopoulos, and G.~Caire, ``On-the-fly
  large-scale channel-gain estimation for massive antenna-array base
  stations,'' in \emph{2018 IEEE International Conference on Communications
  (ICC)}, 2018, pp. 1--6, doi:10.1109/ICC.2018.8422419.

\bibitem{9374476}
A.~Fengler, S.~Haghighatshoar, P.~Jung, and G.~Caire, ``Non-{Bayesian} activity
  detection, large-scale fading coefficient estimation, and unsourced random
  access with a massive {MIMO} receiver,'' \emph{IEEE Transactions on
  Information Theory}, vol.~67, no.~5, pp. 2925--2951, 2021,
  doi:10.1109/TIT.2021.3065291.

\bibitem{Rician1}
F.~Tian, X.~Chen, L.~Liu, and D.~W.~K. Ng, ``Massive unsourced random access
  over {Rician} fading channels: Design, analysis, and optimization,''
  \emph{IEEE Internet of Things Journal}, vol.~9, no.~18, pp. 17\,675--17\,688,
  2022, doi:10.1109/JIOT.2022.3155670.

\bibitem{Rician2}
W.~Liu, Y.~Cui, F.~Yang, L.~Ding, and J.~Sun, ``{MLE}-based device activity
  detection for grant-free massive access under {Rician} fading,'' in
  \emph{2022 IEEE 23rd International Workshop on Signal Processing Advances in
  Wireless Communication (SPAWC)}, 2022, pp. 1--5,
  doi:10.1109/SPAWC51304.2022.9833944.

\bibitem{gmmvamp}
M.~Ke, Z.~Gao, Y.~Wu, X.~Gao, and R.~Schober, ``Compressive sensing-based
  adaptive active user detection and channel estimation: Massive access meets
  massive {MIMO},'' \emph{IEEE Transactions on Signal Processing}, vol.~68, pp.
  764--779, 2020, doi:10.1109/TSP.2020.2967175.

\bibitem{thabane2004matrix}
L.~Thabane and M.~Safiul~Haq, ``On the matrix-variate generalized hyperbolic
  distribution and its {Bayesian} applications,'' \emph{Statistics}, vol.~38,
  no.~6, pp. 511--526, 2004.

\bibitem{cheng2022towards}
L.~Cheng, Z.~Chen, Q.~Shi, Y.-C. Wu, and S.~Theodoridis, ``Towards flexible
  sparsity-aware modeling: Automatic tensor rank learning using the generalized
  hyperbolic prior,'' \emph{IEEE Transactions on Signal Processing}, vol.~70,
  pp. 1834--1849, 2022, doi:10.1109/TSP.2022.3164200.

\bibitem{covchen2022}
Z.~Chen, F.~Sohrabi, Y.-F. Liu, and W.~Yu, ``Phase transition analysis for
  covariance-based massive random access with massive {MIMO},'' \emph{IEEE
  Transactions on Information Theory}, vol.~68, no.~3, pp. 1696--1715, 2022,
  doi:10.1109/TIT.2021.3132397.

\bibitem{bishop2006pattern}
C.~M. Bishop and N.~M. Nasrabadi, \emph{Pattern recognition and machine
  learning}.\hskip 1em plus 0.5em minus 0.4em\relax Springer, 2006, vol.~4,
  no.~4.

\bibitem{etaprior}
S.~D. Babacan, S.~Nakajima, and M.~N. Do, ``{Bayesian} group-sparse modeling
  and variational inference,'' \emph{IEEE Transactions on Signal Processing},
  vol.~62, no.~11, pp. 2906--2921, 2014, doi:10.1109/TSP.2014.2319775.

\bibitem{beal2003variational}
M.~J. Beal, \emph{Variational algorithms for approximate {Bayesian}
  inference}.\hskip 1em plus 0.5em minus 0.4em\relax University of London,
  University College London (United Kingdom), 2003.

\bibitem{boyd2004convex}
S.~Boyd, S.~P. Boyd, and L.~Vandenberghe, \emph{Convex optimization}.\hskip 1em
  plus 0.5em minus 0.4em\relax Cambridge university press, 2004.

\bibitem{MLMurphy}
K.~P. Murphy, \emph{Machine learning: a probabilistic perspective}, Cambridge,
  MA, 2012.

\end{thebibliography}
\bibliographystyle{IEEEtran}
\begin{IEEEbiography}
	[{\includegraphics[width=1in,height=1.25in,clip,keepaspectratio]{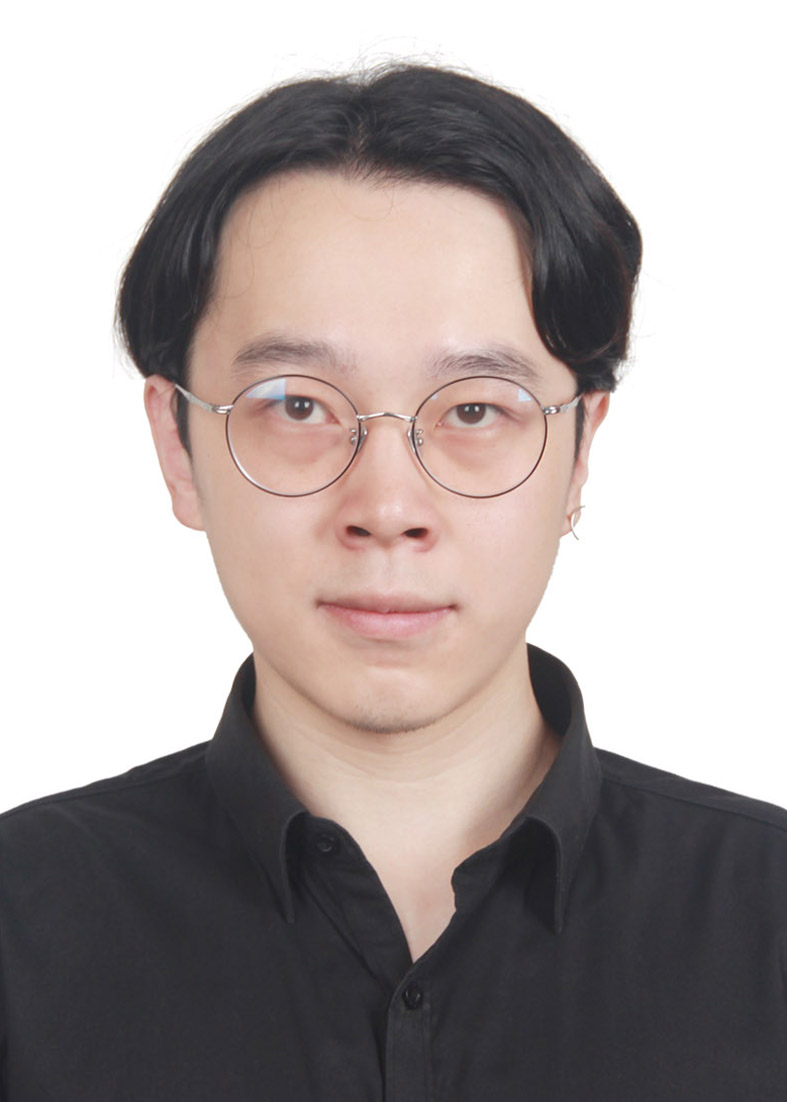}}]  {Hao Zhang} received the B.Eng. degree from Zhejiang University, Hangzhou, China, in 2019. He is currently working toward the Ph.D. degree with the Department of Electrical and Electronic Engineering, The University of Hong Kong, Hong Kong. His research interests include Bayesian machine learning, and wireless communication.
\end{IEEEbiography}
\vspace{-400pt}
\begin{IEEEbiography}
	[{\includegraphics[width=1in,height=1.25in,clip,keepaspectratio]{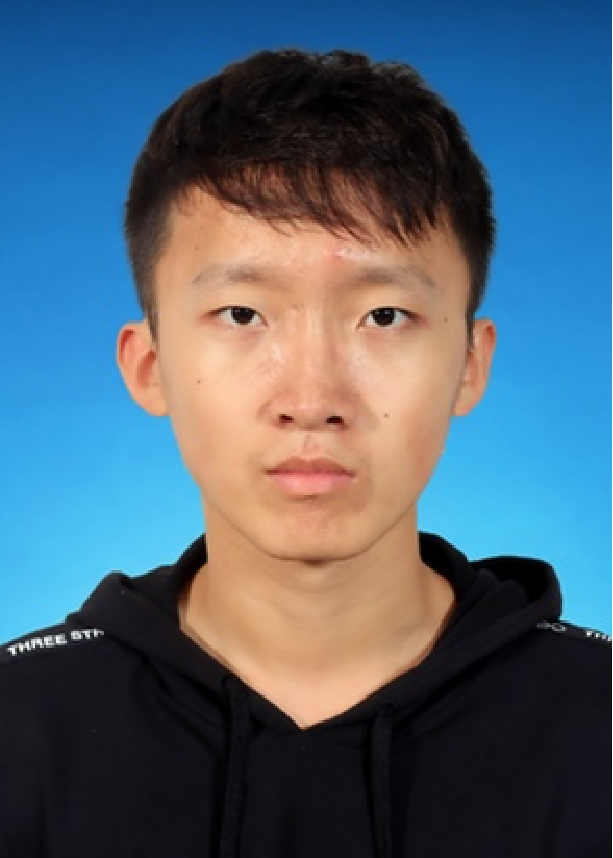}}]  {Qingfeng Lin} received the B.Eng. degree in communication engineering and the M.Eng. degree in information and communication engineering from the Harbin Institute of Technology, Harbin, China, in 2018 and in 2020, respectively. He is currently working toward the Ph.D. degree with the Department of Electrical and Electronic Engineering, The University of Hong Kong, Hong Kong. His research interests include large-scale optimization, machine learning in wireless communications.
\end{IEEEbiography}
\vspace{-400pt}
\begin{IEEEbiography}
	[{\includegraphics[width=1in,height=1.25in,clip,keepaspectratio]{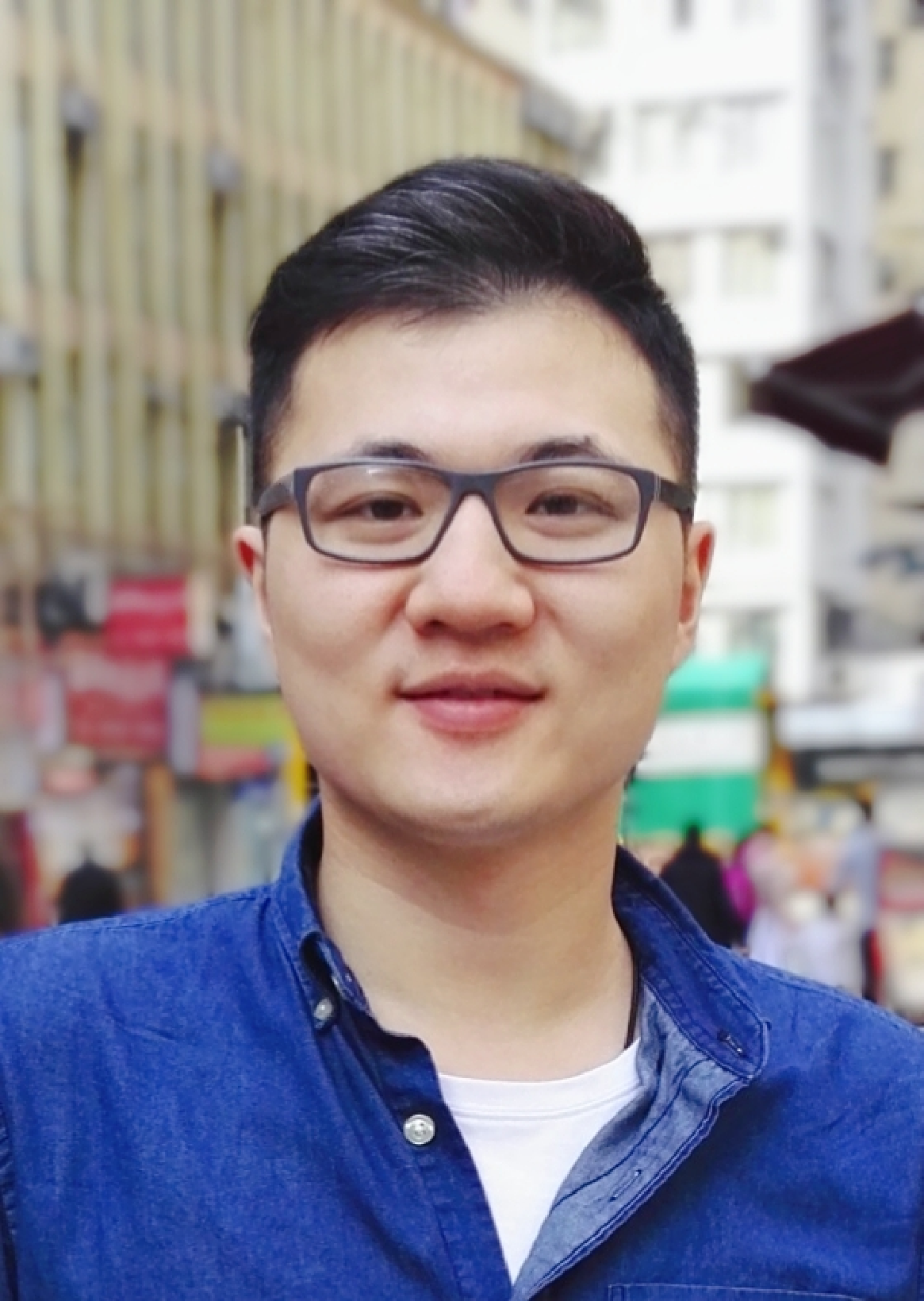}}]  {Yang Li} (Member, IEEE) received the B.E. and M.E. degrees in electronics engineering from Beihang University (BUAA), Beijing, China, in 2012 and 2015, respectively, and the Ph.D. degree from the Department of Electrical and Electronic Engineering, The University of Hong Kong (HKU), in 2019. From 2019 to 2020, he has been a Senior Research Engineer with Huawei Noah’s Ark Laboratory. He is currently a Research Scientist with the Shenzhen Research Institute of Big Data. His research interests include radio resource management, learning to optimize, and large-scale optimization. He is the Winner of the 2020 Innovation Pioneer Award of Huawei.
\end{IEEEbiography}
\newpage
\begin{IEEEbiography}
	[{\includegraphics[width=1in,height=1.25in,clip,keepaspectratio]{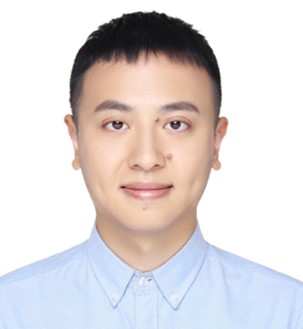}}]  {Lei Cheng} is currently Assistant Professor (ZJU Young Professor) with the College of Information Science and Electronic Engineering, Zhejiang University, Hangzhou, China. He received the B.Eng. degree from Zhejiang University in 2013, and the Ph.D. degree from The University of Hong Kong in 2018. He was a Research Scientist in Shenzhen Research Institute of Big Data, The Chinese University of Hong Kong, Shenzhen, from 2018 to 2021. He is the co-author of the book “Bayesian Tensor Decomposition for Signal Processing and Machine Learning: Modeling, Tuning-Free Algorithms, and Applications”, Springer, 2023. He was a Tutorial Speaker in ICASSP 2023. His research interests are in Bayesian machine learning for tensor data analytics, and interpretable machine learning for information systems.
\end{IEEEbiography}
\vspace{-400pt}
\begin{IEEEbiography}
	[{\includegraphics[width=1in,height=1.25in,clip,keepaspectratio]{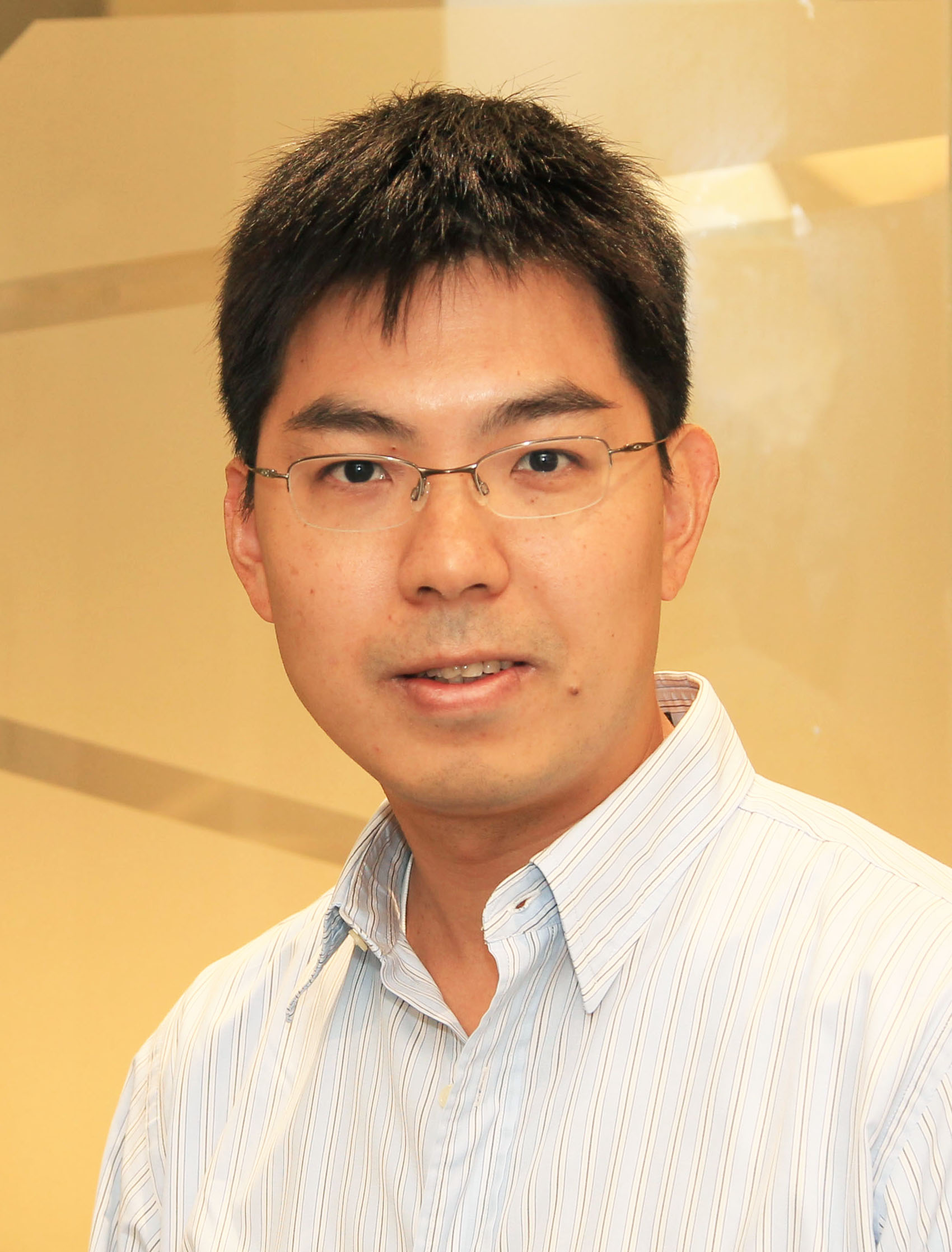}}]  {Yik-Chung Wu} received the B.Eng. (EEE) and M.Phil. degrees from The University of Hong Kong (HKU), in 1998 and 2001, respectively, and the Ph.D. degree from Texas A\&M University, College Station, in 2005. From 2005 to 2006, he was with the Thomson Corporate Research, Princeton, NJ, as a member of Technical Staff. Since 2006, he has been with HKU, currently as an Associate Professor. He was a Visiting Scholar at Princeton University in 2015 and 2017. His research interests include general areas of communication systems, signal processing, and machine learning. He received four best paper awards in international conferences, with the most recent one from IEEE International Conference on Communications (ICC) 2020. He served as an Editor for the IEEE COMMUNICATIONS LETTERS and the IEEE TRANSACTIONS ON COMMUNICATIONS. He is currently a Senior Area Editor for IEEE TRANSACTIONS ON SIGNAL PROCESSING, an Associate Editor for IEEE WIRELESS COMMUNICATION LETTERS and an Editor of Journal of Communications and Networks.
\end{IEEEbiography}

\end{document}